\documentclass[5p,authoryear,10pt]{elsarticle}  

\usepackage{graphics}
\usepackage{graphicx}

\usepackage{amssymb}

\usepackage{microtype}

\journal{Neural Networks}

\begin{document}

\begin{frontmatter}



\title{A Neuromorphic VLSI Design for Spike Timing \\ 
and Rate Based Synaptic Plasticity}

\author[first]{Mostafa Rahimi Azghadi\corref{cor1}}
\ead{mostafa@eleceng.adelaide.edu.au}

\author[first]{Said Al-Sarawi\corref{cor1}}
\ead{alsarawi@eleceng.adelaide.edu.au}

\author[first]{Derek Abbott}
\ead{dabbott@eleceng.adelaide.edu.au}

\author[first]{Nicolangelo Iannella\corref{cor1}} 
\ead{iannella@eleceng.adelaide.edu.au}

\address[first]{School of Electrical and Electronic Engineering,\\
The University of Adelaide, Adelaide, SA 5005, Australia}
\cortext[cor1]{To whom correspondence should be addressed to}



\begin{abstract}
Triplet-based Spike Timing Dependent Plasticity (TSTDP) is a powerful synaptic plasticity rule that acts beyond conventional pair-based STDP (PSTDP). Here, the TSTDP is capable of reproducing the outcomes from a variety of biological experiments, while the PSTDP rule fails to reproduce them. Additionally, it has been shown that the behaviour inherent to the spike rate-based Bienenstock-Cooper-Munro (BCM) synaptic plasticity rule can also emerge from the TSTDP rule. This paper proposes an analog implementation of the TSTDP rule. The proposed VLSI circuit has been designed using the AMS $0.35~\mu$m CMOS process and has been simulated using design kits for Synopsys and Cadence tools. Simulation results demonstrate how well the proposed circuit can alter synaptic weights according to the timing difference amongst a set of different patterns of spikes. Furthermore, the circuit is shown to give rise to a BCM-like learning rule, which is a rate-based rule. To mimic implementation environment, a 1000 run Monte Carlo (MC) analysis was conducted on the proposed circuit. The presented MC simulation analysis and the simulation result from fine-tuned circuits show that, it is possible to mitigate the effect of process variations in the proof of concept circuit, however, a practical variation aware design technique is required to promise a high circuit performance in a large scale neural network. We believe that the proposed design can play a significant role in future VLSI implementations of both spike timing and rate based neuromorphic learning systems.
\end{abstract}


\end{frontmatter}


\section{Introduction}
\label{Intro}

The underlying mechanisms and processes responsible for learning and long-term memory in the brain has remained an important yet strongly debated subject for researchers in various fields ranging from neurophysiology through to neuromorphic engineering. It is widely believed that processes responsible for synaptic plasticity provide key mechanisms underlying learning and memory in the brain \cite{Song2000,Pfister2006,Sjostrom2001,Wang2005}. Researchers from various fields, including biology, neurophysiology, and engineering over the last fifty years have attempted to explore, describe and understand the processes of synaptic plasticity leading to learning and memory. Engineers typically attempt to emulate and/or mimic biological systems to various degrees of detail and description. Neuromorphic engineers have been working concurrently,  for the last two decades, with neurobiologists to implement various neuron models and learning rules in electronic circuits~\cite{Mead1989,Bamford2012}. These electronic circuits lead to a very high degree of parallelism, as well as ultra dense physical realizations, which are major advantages for implementing practical neural networks. There are various types of neurons implemented as electronic circuits that can be found in the literature e.g.~\cite{Simoni2004,Farquhar2005,Indiveri2006,Indiveri2011}. Furthermore, there exists various circuit implementations for various learning rules, specifically spike timing-dependent plasticity (STDP)~\cite{Bofill-I-Petit2004,Indiveri2006,Tanaka2009,Mayr2010,Meng2011,Rachmuth2011,Ramakrishnan2011,Bamford2012}.

Neurophysiological experiments have illustrated that plastic changes to synapses can occur via spike-timing, varying the frequency of inputs to the neuron, or changes to internal concentration of calcium in the neuron's spine apparatus~\cite{Bi1998,Sjostrom2001,Wang2005}. Many theoretical and experimental studies have focused on studying changes to synaptic strength caused by STDP~\cite{Gerstner1996,Bi1998,Song2000,Froemke2002,Wang2005,Pfister2006,Iannella2006,Iannella2010}. At the same time, there have been attempts at translating such rules to VLSI circuit implementations~\cite{Bofill-I-Petit2004,Indiveri2006,Tanaka2009,Rachmuth2011,RahimiAzghadi2011,RahimiAzghadi2012a,RahimiAzghadi2012}.~These attempts represent the crucial technological steps in developing smart VLSI chips with adaptive capabilities similar to that of the mammalian brain. The long term aim is to have VLSI circuits that can learn to adapt to changes and result in modifying their functionality to improve their performance. The realization of such adaptive VLSI circuits will have widely varying applications ranging from artificial eyes through to improved autonomous systems.

The main contribution of this study is to introduce a new synaptic analog circuit which possesses some critical capabilities that have not been demonstrated in previous VLSI implementations. The proposed circuit not only can replicate known outcomes of STDP, including the effects of 
input frequency, but also  it is capable of mimicking BCM-like behaviour~\cite{Bienenstock1982}. The proposed circuit captures important aspects of both timing- and rate-based synaptic plasticity that is of great interest for researchers in the field of neuromorphic engineering, specifically to those who are involved in experiments dealing with learning and memory {\it in-silico}. 

The paper is organized as follows. In Section~\ref{Syn}, a brief overview of timing- and rate-based synaptic plasticity rules is given. Section~\ref{VLSI} briefly reviews and discusses some previous VLSI implementations of different synaptic plasticity rules and their capabilities. In Section~\ref{prop}, a description of the proposed circuit operation is given. Section~\ref{Sim} is dedicated to simulation results where we illustrate the capabilities of our proposed circuit. Section~\ref{mis} discusses and describes the effects of process variation and transistor mismatch on the proposed design, and suggests a tuning mechanism to overcome the performance degradation in the presence of physical variations. Section~\ref{conc}, provides both a discussion of current trends and future outlooks including concluding remarks.

\section{Synaptic Plasticity Rules}
\label{Syn}
Synapses are specialised structures that allow either chemical or electrical signals to pass from the pre-synaptic to the target post-synaptic neuron with an associated synaptic strength or efficacy. The weight or efficacy of a synapse can be typically altered through the action of neural activity, where the dynamics and time course of synaptic change can be described by various prescriptions called synaptic plasticity rules. These rules dictate the direction of synaptic weight modification, that is whether a synaptic depression or potentiation occurs ultimately leading to changes in the responses of spiking neurons. These changes are associated to different dynamical states of the cortical network. Such synaptic dynamics is believed to be the essential requirement for learning. Synaptic plasticity rules describe (molecular) processes which induce weight changes according to their various interpretations of pre-synaptic and post-synaptic neurons firing activities and current conditions. Some rules consider the precise timing of synaptic inputs and the generation of action potentials to induce synaptic weight changes, while others may take the rate or average of pre- and post-synaptic neurons firing activities into account for altering the strengths of synapses. Other hybrid rules utilise both spike rate and spike timing simultaneously to vary the synaptic weight. Furthermore, some rules act in a nonlinear weight-dependent manner to determine the future state of the post-synaptic neuron. All mentioned rules are concisely reviewed in the following subsections.

\subsection{Timing-based Synaptic Plasticity Rules}
\subsubsection{Pair-based STDP} 
The pair-based rule is the classical description of STDP, which has been widely used in various studies as well as several VLSI implementations~\cite{Bofill-I-Petit2004,Cameron2005,Indiveri2006,Tanaka2009,Mayr2010,Meng2011,Bamford2012}. The original rule expressed by Eq.~\ref{eq:stdppair} is a mathematical representation of the pair-based STDP rule~\cite{Song2000}.
\begin{equation} \label{eq:stdppair}
\Delta w = \left\{\begin{array}{rl}
\Delta w^+=A^+e^{(\frac{-\Delta t}{\tau_+})} & \mbox{if}~\Delta t>0 \\
\Delta w^-=-A^-e^{(\frac{\Delta t}{\tau_-})} & \mbox{if}~\Delta t \leq 0,
\end{array} \right.
\end{equation}
where $\Delta t=t_{\rm post}-t_{\rm pre}$ is the timing difference between a single pair of pre- and post-synaptic spikes. According to this model, the synaptic weight will be potentiated if a pre-synaptic spike arrives in a specified time window ($\tau_+$) before the occurrence of a post-synaptic spike. Analogously, depression will occur if a pre-synaptic spike occurs after the post-synaptic spike. The amount of potentiation/depression will be determined as a function of the timing difference between pre- and post-synaptic spikes, their temporal order, and their relevant amplitude parameters ($A^+$ and $A^-$).

\subsubsection{Triplet-based STDP} \label{sec:triplet stdp}
In this model of synaptic plasticity, changes to synaptic weight are based on the timings of triplet combination of spikes~\cite{Pfister2006}. This rule uses higher order temporal patterns of spikes to modify the weights of synapses and is called triplet-based spike timing-dependent plasticity (TSTDP). A mathematical representation of this learning rule is given by 
\begin{equation}\label{eq:stdptrip}
\Delta w = \left\{ \begin{array}{rl}
\Delta w^+=e^{(\frac{-\Delta t_1}{\tau_+})}\Big(A_2^+ +A_3^+e^{(\frac{-\Delta t_2}{\tau_y})}\Big) \\
\Delta w^-=-e^{(\frac{\Delta t_1}{\tau_-})}\Big(A_2^- +A_3^-e^{(\frac{-\Delta t_3}{\tau_x})}\Big),
\end{array} \right.
\end{equation}
where $\Delta w=\Delta w^+$ for $t=t_{\rm post}$ and if $t=t_{\rm pre}$ then the weight change is $\Delta w=\Delta w^-$.~$A_2^+$, $A_2^-$, $A_3^+$ and $A_3^-$ are potentiation and depression amplitude parameters, $\Delta t_1=t_{\rm post(n)}-t_{\rm pre(n)}$, $\Delta t_2=t_{\rm post(n)}-t_{\rm post(n-1)}-\epsilon$ and $\Delta t_3=t_{\rm pre(n)}-t_{\rm pre(n-1)}-\epsilon$, are the time differences between combinations of pre- and post-synaptic spikes. Here,~$\epsilon$ is a small positive constant which ensures that the weight update uses the correct values occurring just before the pre or post-synaptic spike of interest, and finally $\tau_-$, $\tau_+$, $\tau_x$ and $\tau_y$ are time constants~\cite{Pfister2006}. 
Prior to this TSTDP model, there was another rule proposed by~\cite{Froemke2002} which considers higher order temporal patterns (quadruplets) of spikes to induce synaptic modification. Both of these  rules tend to explore the impact of higher order spike patterns on synaptic plasticity. In this study, the proposed analog circuit aims to mimic the model presented in Eq.~\ref{eq:stdptrip}.

Theoretically, TSTDP rules were proposed to overcome deficiencies in the traditional pair-based STDP in being unable to reproduce the experimental outcomes of various physiological experiments like the data presented in~\cite{Sjostrom2001,Wang2005}. The main advantage of synaptic plasticity rules based upon higher order spike patterns over pair-based rules is the fact that contributions to the overall change in efficacy of traditional additive pair-based rules is essentially linear, while for higher order rules the underlying potentiation and depression contributions do not sum linearly. This fact was reported by~\cite{Froemke2002} as they showed that there are non-linear interactions occurring between consecutive spikes during the presentation of higher order spike patterns. It is this underlying non-linearity, which is captured in such higher order spike-based STDP rules (but is clearly lacking in pair-based STDP) is believed to accurately model such nonlinear interaction among spikes. 

\begin{figure*}[!t]
\centering 
\includegraphics[width=1\textwidth]{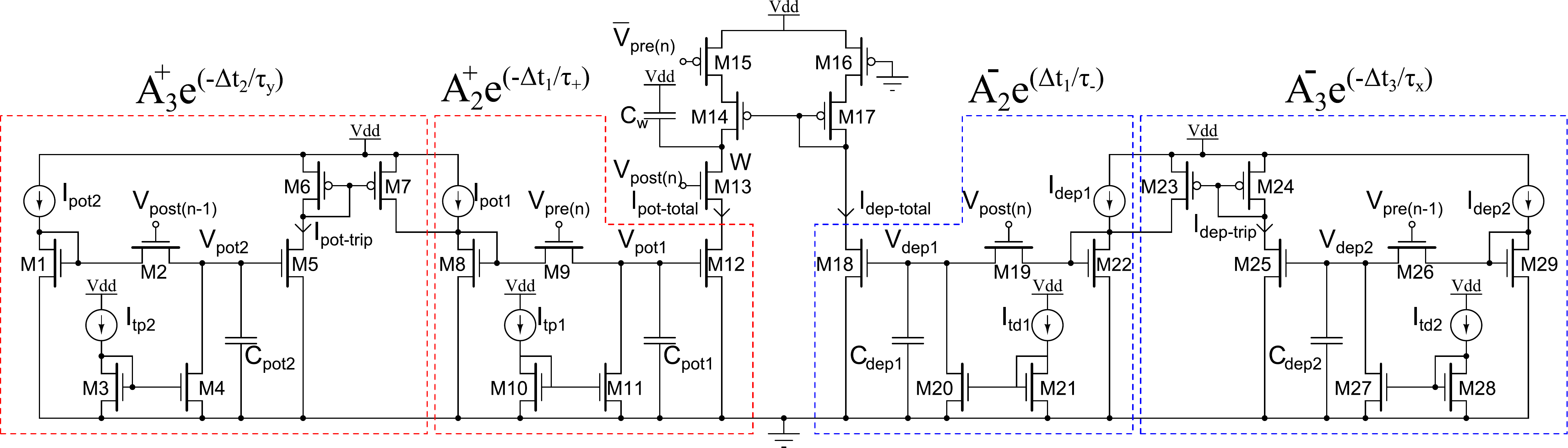}
  \caption{Proposed circuit for the full triplet-based STDP rule. Each part of the circuit aims to implement one part of the rule mentioned in Eq.~\ref{eq:stdptrip}. There are two potentiation parts which are shown in red dashed boxes and two depression parts that are presented in blue dashed boxes. The analytical term corresponding to each part of the circuit is depicted on each box also.}\label{fig:t-stdp}
\end{figure*} 

\subsection{Rate-based Synaptic Plasticity Rules}
As the rate of spikes is one of the major characteristics of a spike train, many researchers over the years have considered this feature as an essential cause of synaptic plasticity ~\cite{Dayan2001}. Two well-known synaptic plasticity rules, which operate according to the rate of pre- and post-synaptic action potentials, are (i) the one proposed by~\cite{Oja1982} and (ii) the experimentally verified Bienenstock-Cooper-Munro (BCM) rule originally proposed by~\cite{Bienenstock1982}. The BCM rule is capable of both Long Term Potentiation (LTP) and Long Term Depression (LTD) whereby the relative proportions of synaptic potentiation and depression can be controlled by a sliding threshold, which also overcomes issues of positive feedback ~\cite{Bienenstock1982}. It will be shown later that although the proposed circuit is an implementation based upon TSTDP, it is also able to demonstrate analogous behaviour to the BCM rule.

According to the BCM rule, synaptic weight change depends linearly on the pre-synaptic, but non-linearly on the post-synaptic activities~\cite{Cooper2004}. Analytically, the triplet STDP rule can show an analogous behaviour of its weight change profile to the BCM rule if the pre- and post-synaptic spike trains are assumed to be Poisson spike trains with the rates $\rho_{\rm pre}$ and $\rho_{\rm post}$, respectively~\cite{Pfister2006}. Traditionally, the mathematical model for the BCM learning rule can be generally represented by
\begin{equation}\label{eq:bcm}
\frac{\Delta w}{\Delta t} = \rho_{\rm pre} \phi(\rho_{\rm post},\theta),
\end{equation}
where $\theta$ is a constant which represents a threshold where the sign of plasticity can change and $\phi$ is a function where $\phi(\rho_{\rm post}<\theta,\theta)<0$ gives rise to a decrease in the synaptic weight (depression), and when $\phi(\rho_{\rm post}>\theta,\theta)>0$, they will be increased (potentiation), and in the case where $\phi(0,\theta)=0$, there will be no change in synaptic weight~\cite{Pfister2006,Cooper2004}.

\subsection{Hybrid Synaptic Plasticity Rules}
There also exists other classes of synaptic plasticity rules which do not fall in either of the groups mentioned above.~We call them hybrid rules as they usually employ a combination of the rate and the timing of action potentials, and sometimes a weight-dependence underlying synaptic weight changes.~One instance of this is Spike Driven Synaptic Plasticity (SDSP) that has been proposed in~\cite{Brader2007}. Note that, SDSP uses both the timing of the pre-synaptic and the rate of the post-synaptic action potentials to modify synaptic weights. Another instance is a similar rule proposed by~\cite{Clopath2010}, which exploits the correlation between pre-synaptic spike arrival times and the voltage of post-synaptic neuron, rather than its time of action potential generation in a phenomenological model of synaptic plasticity. This rule has been shown to account for several biological tasks similar to the TSTDP model proposed by~\cite{Pfister2006}. Additionally another rule that employs a fairly similar method of synaptic modification to the first and second rules mentioned earlier, is proposed by~\cite{Mayr2010a}. Identically, this rule acts based on the transmission profile of the pre-synaptic neuron spikes as well as the post-synaptic membrane voltage.

These above-mentioned rules (including TSTDP) have been proposed in an attempt to develop learning rules which not only can reproduce known experimental outcomes, but also provide a unified framework for synaptic change. So they would lead to more powerful synaptic modification rules, which in turn result in improved learning algorithms. As these rules are being proposed, neuromorphic engineers have begun translating the rules for synaptic plasticity into VLSI implementations. In the following section, a brief overview of the previous VLSI implementations for the above mentioned rules is presented. Next, in Section~\ref{prop}, the proposed circuit is presented and its internal operation is explained, followed by simulation results and rigorous analysis of our newly proposed circuit in the following sections.

\section{VLSI Implementations of Synaptic Plasticity}
\label{VLSI}
Various VLSI implementations of synaptic plasticity rules can be found in the literature.~A large group of these implementations, including those presented in~\cite{Bofill-I-Petit2004,Cameron2005,Indiveri2006,Schemmel2006,Koickal2007,Tanaka2009,Ramakrishnan2011}, try to mimic just the LTP and LTD behaviour of the PSTDP rule and produce a learning window that fits the original experimental data reported in~\cite{Bi1998}. 
There are some other implementations that try to implement other learning rules mentioned in Section~\ref{Syn}.~\cite{Mitra2009} implemented the hybrid learning rule presented in~\cite{Brader2007} and utilized this implementation in a neuromorphic system for real-time classification of complex patterns. Also,~\cite{Mayr2010} presented a circuit that works according to their proposed hybrid plasticity rule presented in~\cite{Mayr2010a}. Furthermore, there are some other VLSI implementations of synaptic plasticity rules, e.g.~the circuit presented by~\cite{Rachmuth2011} that is capable of emulating both PSTDP and the BCM.
To the best of our knowledge, a design or implementation of TSTDP does not exist in the literature other than our initial prototype circuits~\cite{RahimiAzghadi2011,RahimiAzghadi2012}.~In contrast to those circuits, the structure of the presented circuit in this paper is different and it is a close fit to the TSTDP model. As it will be shown later, the new circuit produces a significant fit to reported experimental data and results in much smaller fitting errors compared to previous circuits.

\begin{figure*} [!t]
\centering
  \includegraphics[width=1\textwidth]{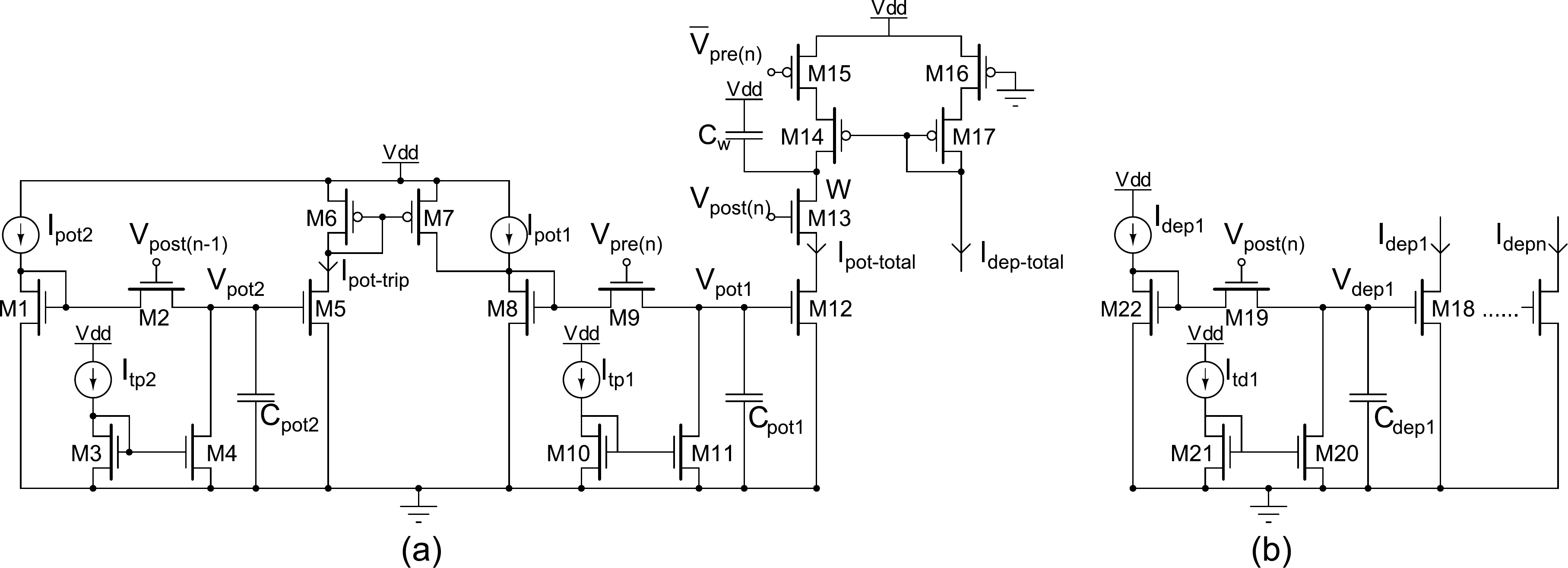}
  \caption{Proposed minimal triplet-based STDP circuit. The synaptic weight is inversely proportional to the value over weight capacitor, $C_{\rm W}$. (a) This part of the circuit brings about potentiation due to pre-post and post-pre-post combinations of spikes. Potentiation means discharging the weight capacitor through M12-M13. This part of the circuit should be replicated for all synapses on all dendrite branches come from various pre-synaptic neurons. (b) This section of the circuit is responsible for depression through charging the weight capacitor. This part needs to be implemented only once per neuron and it can result in area saving as it does not need to be replicated for all synapses.}\label{fig:t-stdp1}
\end{figure*} 

\section{Proposed Circuit for TSTDP Rule}
\label{prop}
This paper proposes a novel VLSI implementation that builds upon a previous study by Bofill-I-Petit and Murray (2004). The new circuit produces a close fit to the outcomes of the TSTDP rule. Fig.~\ref{fig:t-stdp} presents the proposed circuit implementation of the full TSTDP model. In the full TSTDP model, there are eight parameters that can be tuned in the proposed circuit, by controlling eight bias currents as follows: $I_{\rm dep1}$, $I_{\rm pot1}$, $I_{\rm dep2}$ and $I_{\rm pot2}$ represent the amplitude of synaptic weight changes for post-pre ($A_2^-$) and pre-post ($A_2^+$) spike pairs, and pre-post-pre ($A_3^-$) and post-pre-post ($A_3^+$) combinations of spike triplets, respectively. Another control parameter for these amplitude values in the circuit is the pulse width of the spikes, which was kept fixed during all experiments in this paper (1~$\mu$s). In addition to these amplitude parameters, the required time constants in the model for post-pre ($\tau_-$), pre-post ($\tau_+$), pre-post-pre ($\tau_x$) and post-pre-post ($\tau_y$) spike patterns, can be adjusted using $I_{td1}$, $I_{\rm tp1}$, $I_{\rm td2}$ and $I_{\rm tp2}$ respectively (see Eq.~\ref{eq:stdptrip} and Fig.~\ref{fig:t-stdp}).

The proposed circuit works as follows: upon the arrival of a pre-synaptic pulse, $V_{\rm pre(n)}$, M9 and M15 are switched on. At this time, $I_{\rm pot1}$ can charge the first potentiation capacitor, $C_{\rm pot1}$, through M9 to the voltage of $V_{\rm pot1}$. After finishing $V_{\rm pre(n)}$, $V_{\rm pot1}$ starts decaying linearly through M11 and with a rate proportional to $I_{\rm tp1}$. Now, if a post-synaptic pulse, $V_{\rm post(n)}$ arrives at M13 in the decaying period of $V_{\rm pot1}$, namely when M12 is still active, the weight capacitor, $C_{\rm W}$, will be discharged through M12-M13 transistors and a potentiation occurs because of the arrival of a post-synaptic pulse in the interval of effect of a pre-synaptic spike (pre-post combination of spikes). Additionally, if a post-synaptic spike has arrived at M19, soon before the current pre-synaptic spike at M15, the weight capacitor can be charged through M14-M15 transistors and a depression happens. This depression happens because the present pre-synaptic spike is in the time of effect of a post-synaptic spike (post-pre combination of spikes). The amount of depression depends on $V_{\rm dep1}$, which itself can be tuned by the relevant amplitude parameter $I_{\rm dep1}$. Also, the activation interval of M18 can be modified by changing the related time constant parameter $I_{\rm td1}$. Furthermore, another contribution to depression can occur if a previous pre-synaptic pulse, $V_{\rm pre(n-1)}$, has arrived at M26 soon enough before the current pre-synaptic happens at M15 and also before a post-synaptic pulse happens at M19. In this situation, the weight capacitor can be charged again through M14-M15 by an amount proportional to an effect of both $V_{\rm dep2}$ and $V_{\rm dep1}$, simultaneously. This triplet interaction leads to the required non-linearity mentioned in the triplet learning rule. A similar description holds for the situation when a post-synaptic pulse occurs at M13 and M19 transistors. But this time one depression will take place in the result of charging the weight capacitor up through M14-M15 and because of arriving a post-synaptic spike at M19 before a pre-synaptic spike at M15. Besides, two potentiation events can happen if an appropriate situation is provided to discharge the weight capacitor because of a pre-post or a post-pre-post combination of spikes. 
Note that, in this implementation, the synaptic strength is inversely proportional to the voltage stored on the weight capacitor, $C_{\rm W}$. However, for the sake of simplicity when comparing the achieved results to experimental data, the weights are shown in a consistent way to biological data, i.e.~potentiation with positive strength and depression with negative strength.

Upon examination of the TSTDP expression (Eq.~\ref{eq:stdptrip}), there are four different parts that need to be implemented, in order to satisfy the equation as accurately as possible. The proposed circuit (Fig.~\ref{fig:t-stdp}) is composed of four leaky integrators which are arranged in a way that form the required addition and multiplications in the formula in a simple manner. Furthermore, in order to have the exponential behaviour required for the TSTDP rule, M5, M12, M18 and M25 are biased in the subthreshold region of operation. The most left part of the circuit implements the potentiation triplet component of the rule using a simple leaky integrator and the resulting current produced by this part ($I_{\rm pot-trip}$) is given by

\begin{equation}\label{eq:trip1}
I_{\rm pot-trip}=A_3^+e^{(\frac{-\Delta t_2}{\tau_y})},
\end{equation}
where $I_{\rm pot2}$ represents~$A_3^+$, $I_{\rm tp2}$ can control~$\tau_y$ and finally $\Delta t_2=t_{\rm post(n)}-t_{\rm post(n-1)}-\epsilon$ controlled by M2 and M13. Next, $I_{\rm pot-trip}$ is added up to~$I_{\rm pot1}$ current which represents $A_2^+$ in the TSTDP formula (Eq.~\ref{eq:stdptrip}). Hence, the amount of current going to M8 transistor is given by

\begin{equation}\label{eq:trip2}
I_{\rm M8}=A_2^++A_3^+e^{(\frac{-\Delta t_2}{\tau_y})}.
\end{equation}
This current then goes to the second leaky integrator on the second left box in Fig.~\ref{fig:t-stdp} and will result in $I_{\rm pot-total}$ passing through M12 and M13 and discharging the weight capacitor, $C_{\rm W}$, hence causes a potentiation equal to~$\Delta w^+$. The amount of this current which is in result of the contribution of both triplet and pair-based spike patterns, can be written as 
\begin{equation}\label{eq:trip3}
I_{\rm pot-total}=e^{(\frac{-\Delta t_1}{\tau_+})}\Big(A_2^+ +A_3^+e^{(\frac{-\Delta t_2}{\tau_y})}\Big),
\end{equation}
where $I_{\rm tp1}$ can control~$\tau_+$ and finally~$\Delta t_1=t_{\rm post(n)}-t_{\rm pre(n)}$ is controlled by M9 and M13.

The same approach applies for the depression part of Eq.~\ref{eq:stdptrip}. There are two leaky integrators (the blue boxes in Fig.~\ref{fig:t-stdp}), each one is responsible for building an exponential current and the final current ($I_{\rm dep-total}$) which will be mirrored through M14 and M17 into the weight capacitor and result in charging the weight capacitor and hence depression. This is the full TSTDP circuit which realizes the full-TSTDP rule (Eq.~\ref{eq:stdptrip}). However, according to the analytical calculations and numerical simulations presented in~\cite{Pfister2006}, some parts of the full TSTDP rule may be omitted without a significant effect on the efficiency of the rule when replicating biological experiments. Pfister and Gerstner called these new modified rules, minimal triplet rules.

According to the first minimal TSTDP rule, when generating the biological experiment outcomes for the visual cortex data set presented in~\cite{Sjostrom2001}, the triplet contribution for depression, as well as the pairing contribution of the potentiation parts of~Eq.~\ref{eq:stdptrip} can be dismissed (i.e.~$A_3^-=0$ and $A_2^+=0$) and the outcome will be quite similar to using the full TSTDP rule (Table 3 in~\cite{Pfister2006}).
Furthermore, the second minimal TSTDP rule which considers a zero value for $A_3^-$ (Eq.~\ref{eq:stdptrip}) has quite similar consequences to the full TSTDP rule  and allows reproducing the hippocampal culture data set experimental data presented in~\cite{Wang2005}.

As the rules are simplified, the full TSTDP circuit also can be minimized. This minimization can be performed by removing those parts of the circuit that correspond to the omitted parts from the full TSTDP model. These parts are M23-M29 transistors which can be removed when~$I_{\rm dep2}=0$ (i.e.~$A_3^-=0$). Also~$I_{\rm pot1}$ can be set to zero, as it represents~$A_2^+$ that is not necessary for the first minimal triplet rule. The resulting minimal circuit based on these assumptions is shown in Fig.~\ref{fig:t-stdp1} with two separate parts for potentiation and depression. The potentiation part (a) which is composed of two leaky integrators is responsible for voltage decrements across the weight capacitor (potentiation), in case of pre-post or post-pre-post of spike patterns in the required timing periods. This part receives two inputs backpropagated from the post-synaptic neuron ($V_{\rm post(n-1)}$, and $V_{\rm post(n)}$), and another input forwarded from a pre-synaptic neuron ($V_{\rm pre(n)}$). As there can be several synapses on each post-synaptic neuron, this part of the minimal circuit which receives inputs from different pre-synaptic neurons, needs to be replicated for every synapse. However, the depression part of the minimal circuit, part (b), just receives an input from the post-synaptic neuron and hence can be replicated once per neuron. That is why we represent the potentiation and depression inversely to the charge stored on the weight capacitor. As the number of neurons is significantly lower than the number of synapses, this area saving can result in a significantly smaller area for a large neuromorphic system with TSTDP synapses.~A similar approach was also utilized by~\cite{Bofill-I-Petit2004}.

\section{Simulation Results}
\label{Sim}
The proposed circuit shown in Fig.~\ref{fig:t-stdp1} was simulated using parameters for the 0.35$~\mu$m C35 CMOS process by AMS. All transistors in the design are set to 1.05$~\mu$m wide and 0.7$~\mu$m long. The capacitor values are 10~pF for the weight capacitor and 100~fF for all the capacitors in the leaky integrators. The circuit was simulated in Spectre within Cadence and some optimization has been performed using HSpice and Matlab.
All reported experiments in this paper assume the nearest spike interaction, which considers the interaction of a spike only with its two immediate succeeding and preceding nearest neighbours. Furthermore, in order to facilitate the simulation of the circuits, a scaling approach, which has been used in similar VLSI implementations of synaptic plasticity e.g.~\cite{Schemmel2006,Tanaka2009,Mayr2010}, was adopted, which uses a time scale of microseconds to represent milliseconds, i.e a scaling factor of 1000. However, in all simulation results presented in this paper, the results are scaled back to biological time in order to facilitate comparisons with published data from biological experiments.

In order to validate the functionality of the proposed TSTDP circuit, 12 different patterns of spikes including spike pairs (four patterns), spike triplets (six patterns) and spike quadruplets (two patterns) were utilized. These patterns were applied to the circuit and recorded weight changes were compared to their corresponding experimental data. All simulation results show a good match to their related experimental data. The first and second simulations were performed using two different data sets and for different experimental protocols. The optimization scheme and the data fitting method used here were that of~\cite{Pfister2006}. The required experimental protocols, different sets of data, the data fitting method as well as the achieved simulation results, are explained and presented in the following subsections. Additionally, for the third set of simulations, the proposed circuit was examined for generating weight changes using all six possible spike triplet patterns presented in~\cite{Froemke2002}. Furthermore, the circuit was also used to reproduce the weight changes produced by the rate-based BCM rule under a Poissonian protocol. The achieved results for these two simulations, the triplet and Poissonian protocols are also explained in the following subsections.

\subsection{Experimental Protocols}
\subsubsection{Pairing protocol}\label{subsec:exp-pair}
\begin{figure} [!t]
\centering
  \includegraphics[width=0.50\textwidth]{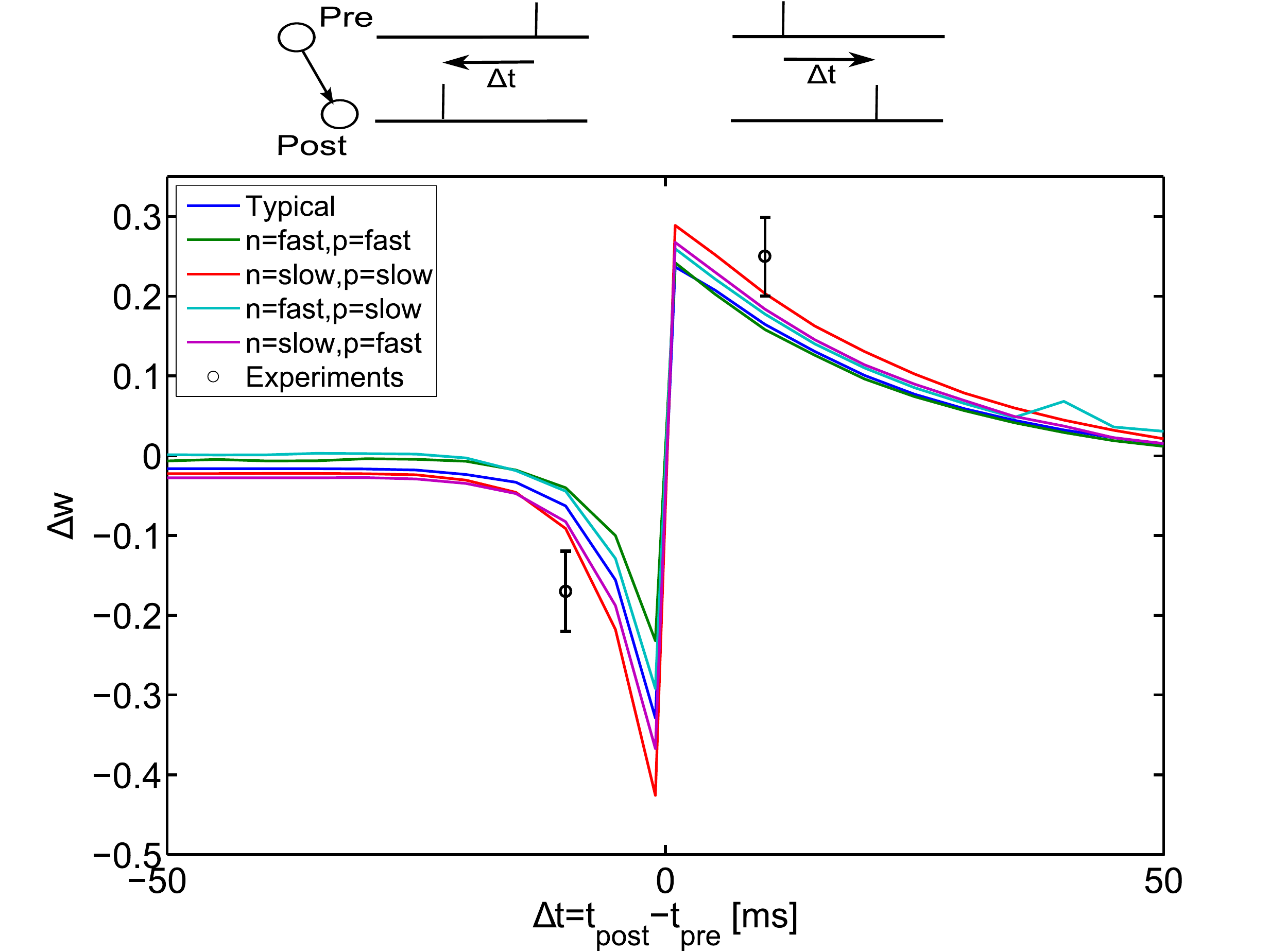}
  \caption{Exponential learning window produced by the proposed minimal TSTDP circuit and based on the pairing protocol for different transistor process corners.~The required bias currents taken for the triplet circuit corresponds to the hippocampal culture data set (Table~\ref{tab:1}). Experimental data and error bars are extracted from~\cite{Wang2005}.}\label{fig:window}
\end{figure}

The pair-based STDP protocol has been extensively used in electrophysiological experiments and simulation studies. In this protocol, 60 pairs of pre- and post-synaptic spikes with a delay of $\Delta t=t_{\rm post}-t_{\rm pre}$ were conducted with a repetition frequency of $\rho$~Hz (in many experiments $\rho=1$~Hz). Fig.~\ref{fig:window} shows that the proposed minimal triplet circuit can reproduce the exponential learning window produced by both PSTDP and TSTDP models, under the conventional pairing protocol described above and adopted in many experiments \cite{Bi1998,Wang2005}. This exponential learning window can also be reproduced using previously described PSTDP circuits e.g.~\cite{Bofill-I-Petit2004}. However, it has been illustrated in~\cite{Sjostrom2001} that altering the pairing repetition frequency affects the total change in weight of the synapse. As it is shown in~\cite{RahimiAzghadi2011,RahimiAzghadi2012}, PSTDP circuits are not capable of reproducing such biological experiments that investigators examine the effect of changes in pairing frequency on synaptic weight. However, Fig.~\ref{fig:pairfreq} illustrates how the proposed TSTDP circuit can readily reproduce these experiments.

\subsubsection{Triplet protocol}\label{subsec:exp-triplet}
There are two types of triplet patterns that are used in the hippocampal experiments, which are also adopted in this paper to compute the prediction error as described in~\cite{Pfister2006}. Both of them consist of 60 triplets of spikes, which are repeated at a given frequency of $\rho=1$~Hz. The first triplet pattern is composed of two pre-synaptic spikes and one post-synaptic spike in a pre-post-pre configuration. As a result, there are two delays between the first pre and the middle post, $\Delta t_1=t_{post}-t_{pre1}$, and between the second pre and the middle post $\Delta t_2=t_{post}-t_{pre2}$. The second triplet pattern is analogous to the first but with two post-synaptic spikes, one before and the other one after a pre-synaptic spike (post-pre-post). Here, timing differences are defined as $\Delta t_1=t_{post1}-t_{pre}$ and $\Delta t_2=t_{post2}-t_{pre}$. Figures~\ref{fig:prepospre} and \ref{fig:posprepos} show how the proposed minimal triplet circuit produces a close fit to the triplet experiments reported in~\cite{Wang2005}.

\subsubsection{Quadruplet protocol}\label{subsec:exp-quadruplet}
This protocol is composed of 60 quadruplets of spikes repeated at frequency of $\rho=1$~Hz.~The quadruplet is composed of either a post-pre pair with a delay of $\Delta t_1=t_{post1}-t_{pre1}<0$ precedes a pre-post pair with a delay of $\Delta t_2=t_{post2}-t_{pre2}>0$ with a time $T>0$, or a pre-post pair with a delay of $\Delta t_2=t_{post2}-t_{pre2}>0$ precedes a post-pre pair with a delay of $\Delta t_1=t_{post1}-t_{pre1}<0$ with a time $T<0$, where $T=(t_{pre2}+t_{post2})/2-(t_{pre1}+t_{post1})/2$. Identical to~\cite{Pfister2006}, in all quadruplet experiments in this paper, $\Delta t$=$-\Delta t_1$=$\Delta t_2$=5~$\mu s$. Fig.~\ref{fig:quad} shows the weight changes produced by the proposed minimal TSTDP circuit under quadruplet protocol conditions.
Note that none of the previously proposed PSTDP circuits is capable of showing such a close fit, neither to triplet, nor to quadruplet experiments ~\cite{RahimiAzghadi2011,RahimiAzghadi2012}.

\begin{figure} [!t]
\centering
  \includegraphics[width=0.50\textwidth]{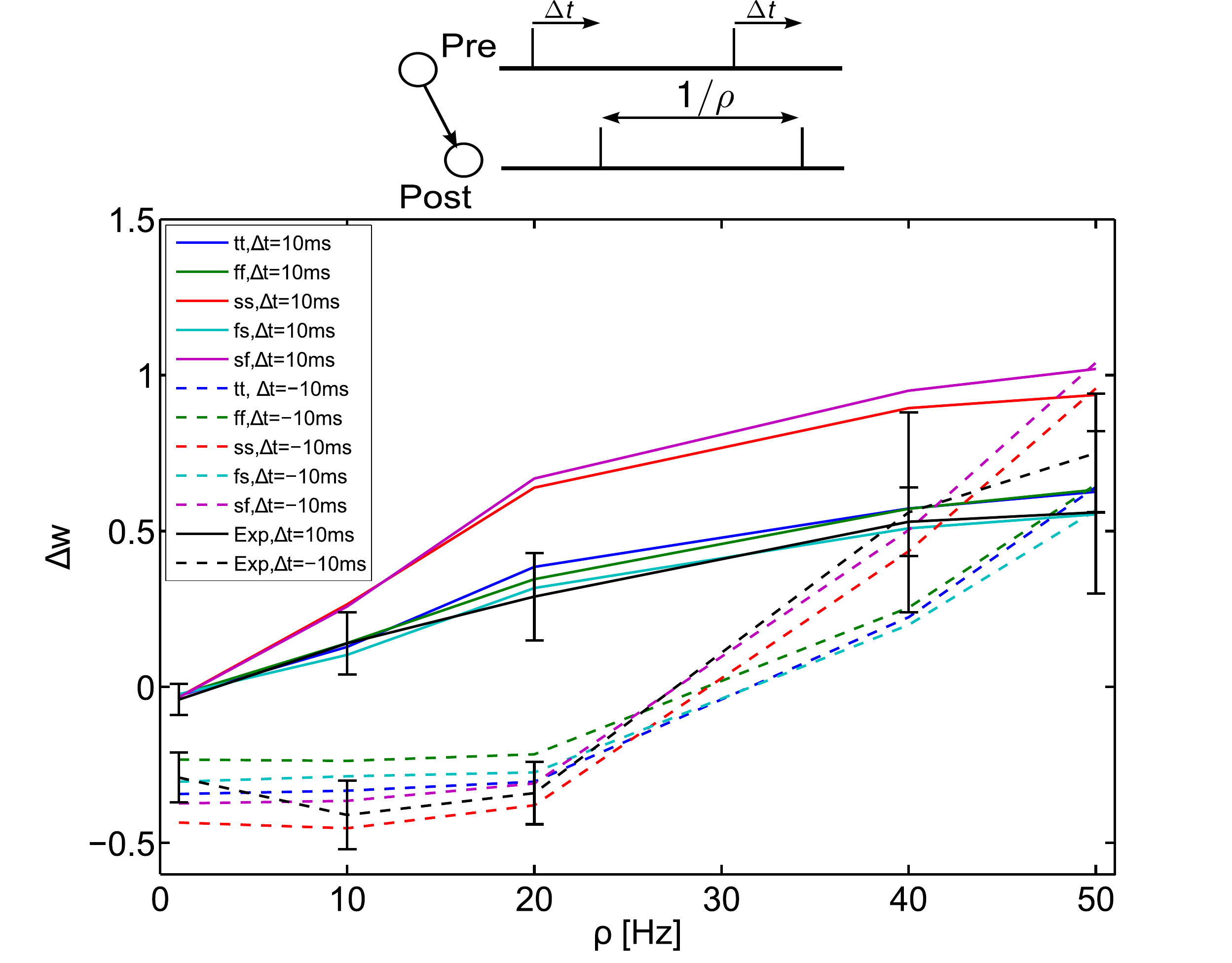}
  \caption{Weight change in a pairing protocol as a function of the pairing frequency~$\rho$ reproduced by the proposed minimal TSTDP circuit for different transistor process corners. Experimental data points and error bars are extracted from~\cite{Sjostrom2001} (no data point at~$\rho = 30$~Hz). The required bias currents taken for the triplet circuit corresponds to the visual cortex data set (Table~\ref{tab:1}).}\label{fig:pairfreq}
\end{figure}

\subsection{Data Sets}\label{subsec:data}
\begin{figure} [!t]
\centering
  \includegraphics[width=0.5\textwidth]{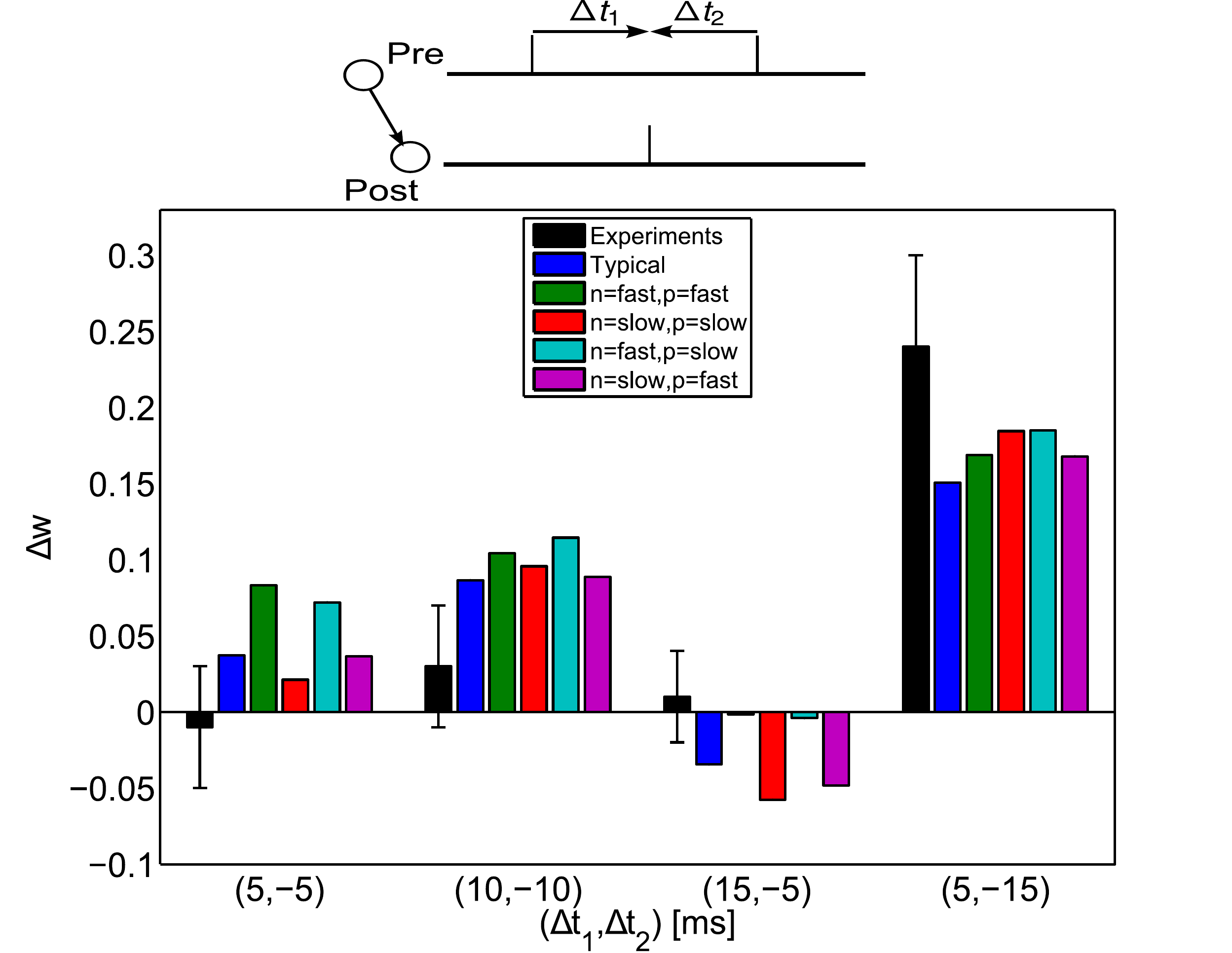}
  \caption{Triplet protocol for the pre-post-pre combination of spikes produced by the proposed minimal TSTDP circuit for different transistor process corners. The required bias currents taken for the triplet circuit corresponds to the hippocampal culture data set (Table~\ref{tab:1}).}\label{fig:prepospre}
\end{figure}

\begin{figure} [!t]
\centering
  \includegraphics[width=0.5\textwidth]{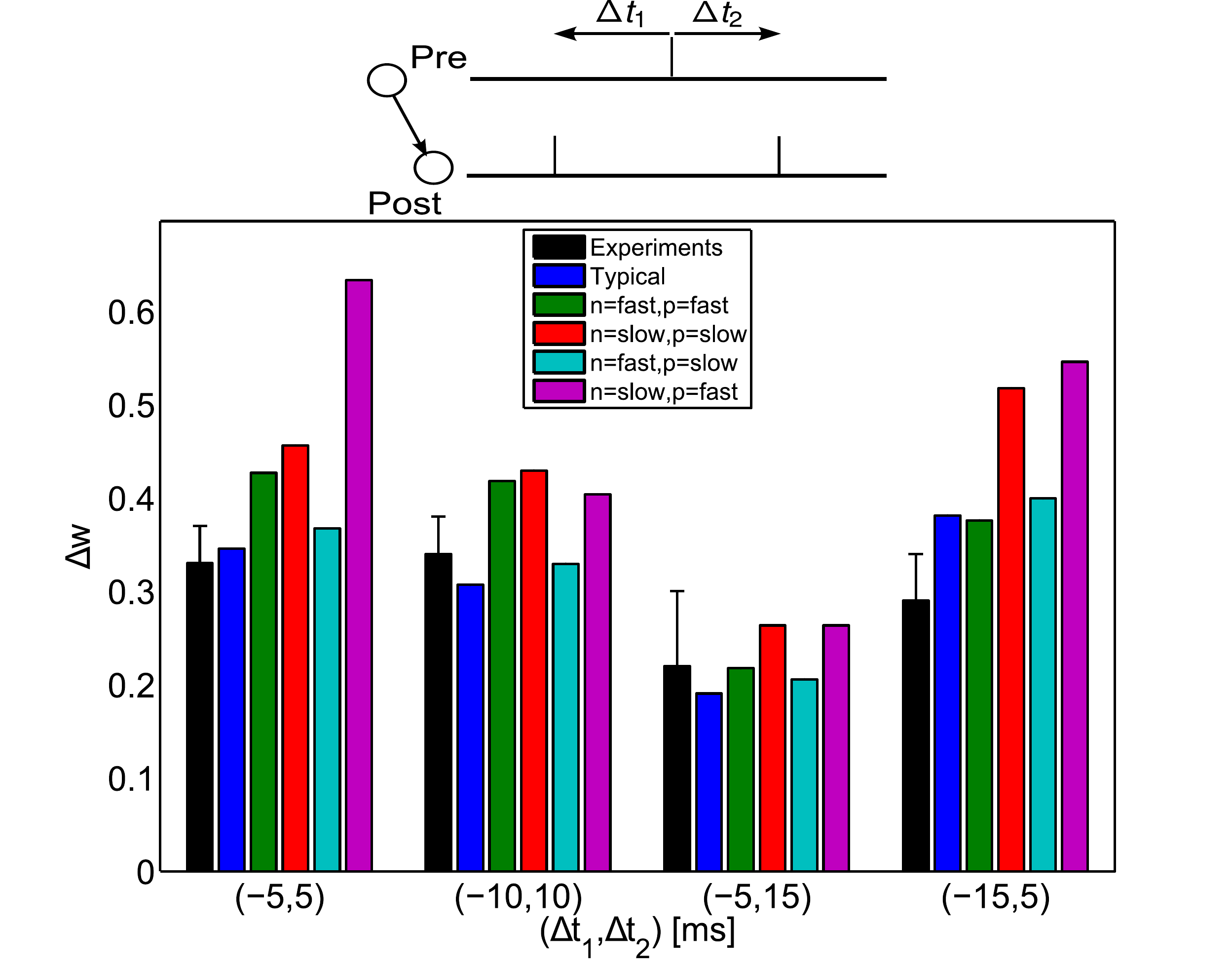}
  \caption{Triplet protocol for the post-pre-post combination of spikes produced by the proposed minimal TSTDP circuit for different transistor process corners. The required bias currents taken for the triplet circuit corresponds to the hippocampal culture data set (Table~\ref{tab:1}).}\label{fig:posprepos}
\end{figure}

The proposed circuit is expected to be capable of reproducing experimental weight changes induced by pairing, triplet and quadruplet protocols in hippocampal cultures reported in~\cite{Wang2005}. It should also be able to reproduce experimental weight changes induced by a pairing protocol and in the presence of spike pairing frequency changes, in the visual cortex presented in~\cite{Sjostrom2001}. In order to check if the proposed circuit is capable of doing so, simulations were conducted using two types of data sets: The first data set originates from experiments on the visual cortex which investigated how altering the repetition frequency of spike pairings affects the overall synaptic weight change. This data set is composed of 10 data points (obtained from Table 1 of~\cite{Pfister2006}) that represents experimental weight change,$~\Delta w$, for two different$~\Delta t$'s, and as a function of the frequency of spike pairs under a pairing protocol in the visual cortex (10 black data points and error bars shown in Fig.~\ref{fig:pairfreq}). The second experimental data set that was utilized originates from hippocampal culture experiments which examines pairing, triplet and quadruplet protocols effects on synaptic weight change. This data set consists of 13 data points obtained from Table 2 of~\cite{Pfister2006} including (i) two data points and error bars for pairing protocol in Fig.~\ref{fig:window}, (ii) three data points and error bars for quadruplet protocol in Fig.~\ref{fig:quad}, and (iii) eight data point and error bars for triplet protocol in Figures~\ref{fig:prepospre} and~\ref{fig:posprepos}. This data set shows the experimental weight change,$~\Delta w$, as a function of the relative spike timing $~\Delta t$, $~\Delta t_1$, $~\Delta t_2$ and $T$ under pairing, triplet and quadruplet protocols in hippocampal culture.

\subsection{Data Fitting Approach}\label{subsec:err}

Identical to \cite{Pfister2006} that test their proposed triplet model simulation results against the experimental data using a Normalized Mean Square Error (NMSE) for each of the data sets, the proposed circuit is verified by comparing its simulation results with the experimental data and ensuring a small NMSE value. The NMSE is calculated using the following equation:

\begin{eqnarray}\label{eq:err}
{\rm NMSE}=\frac{1}{p}\sum_{i=1}^p\left( \frac{\Delta w^i_{\rm exp}-\Delta w^i_{\rm cir}}{\sigma_i}\right)^{2}\hspace{-2mm},\label{eq:3}
\end{eqnarray}
where $\Delta w^i_{\rm exp}$, $\Delta w^i_{\rm cir}$ and $\sigma_i$ are the mean weight change obtained from biological experiments, the weight change obtained from the circuit under consideration, and the standard error mean of $\Delta w^i_{\rm exp}$ for a given data point $i$, respectively; $p$ represents the number of data points in a specified data set (can be 10 or 13).

In order to minimize the resulting NMSEs for the circuit and fit the circuit output to the experimental data, there is a need to adjust the circuit bias parameters and time constants. This is an optimization process of the circuit bias currents which results in reaching a minimum NMSE value and so the closest possible fit to the experimental data.
In the following subsection, the optimization method used to tune the circuit bias currents is introduced.

\subsection{Optimization Method}\label{subsec:opt}
In order to minimize the NMSE function mentioned above and achieve the highest analogy to the experimental data, the circuit bias currents which tunes the required parameters from the model should be optimized as it is the case for both PSTDP and TSTDP model parameters (Eq.~\ref{eq:stdppair} and Eq.~\ref{eq:stdptrip}). For this purpose, Matlab and HSpice were integrated in a way to minimize the NMSE resulted from circuit simulations using the Matlab built-in function {\tt fminsearch}. This function finds the minimum of an unconstrained multi-variable function using a derivative-free simplex search method. Table~\ref{tab:1} demonstrates bias currents achieved from the mentioned optimization method in order to reach the minimum NMSE for the two sets of data: the visual cortex data set and hippocampal culture data set. The minimum obtained NMSEs for the visual cortex and hippocampal data sets are also presented in Table~\ref{tab:1}. These values are consistent with the obtained NMSEs using TSTDP model reported in~\cite{Pfister2006}.

In addition to the above mentioned experiments that have been carried out in~\cite{Pfister2006}, the proposed design has been additionally tested for all possible combination of spike triplets. Applied protocol and more explanation on these experiments are provided in the following subsection.

\begin{figure} [!t]
\centering
  \includegraphics[width=0.5\textwidth]{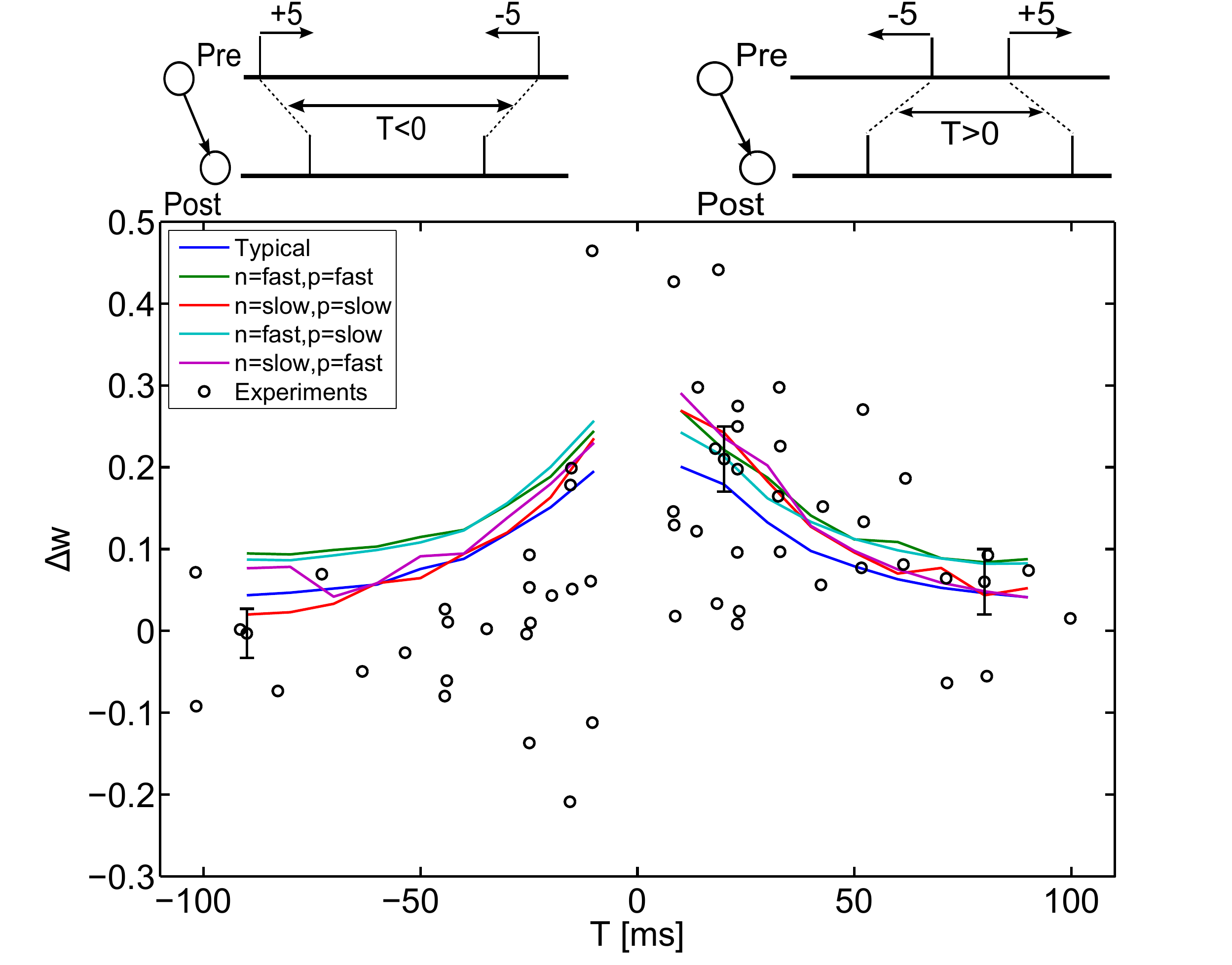}
  \caption{Quadruplet protocol produced by the proposed minimal TSTDP circuit for different transistor process corners. The required bias currents taken for the triplet circuit corresponds to the hippocampal culture data set (Table~\ref{tab:1}). Experimental data points and error bars are after~\cite{Sjostrom2001}.}\label{fig:quad}
\end{figure} 

\subsection{Extra Triplet Patterns}\label{subsec:supp}
Apart from reproducing the behaviour of the TSTDP model proposed by~\cite{Pfister2006}, the proposed circuit is also able to reproduce the observed weight modifications for other combinations (rather than pre-post-pre or post-pre-post) of spikes triplets which have not been explored in~\cite{Pfister2006}, but have been used in another set of multi-spike interaction experiments performed by~\cite{Froemke2002}. In these experiments, six different combinations of spike triplets induce synaptic weight changes. These changes in~\cite{Froemke2002} have been calculated according to a suppressive model described as

\begin{equation}
\Delta W_{ij}=\epsilon_i^{\rm pre} \epsilon_j^{\rm post} F(\Delta t_{ij}),
\end{equation}
where $\Delta w_{ij}$ is the synaptic weight change due to the $i^{\rm th}$ pre-synaptic spike and the $j^{\rm th}$ post-synaptic spike, $\epsilon_i=1-e^{-(t_i-t_{(i-1)})/\tau_s}$ is the efficacy of~$i^{\rm th}$ spike and~$\tau_s$ is the suppression constant. In addition $F(\Delta t_{ij})$ is defined in a similar way as defined by Eq.~\ref{eq:stdppair}.

\begin{figure} [!t]
\centering
  \includegraphics[width=0.5\textwidth]{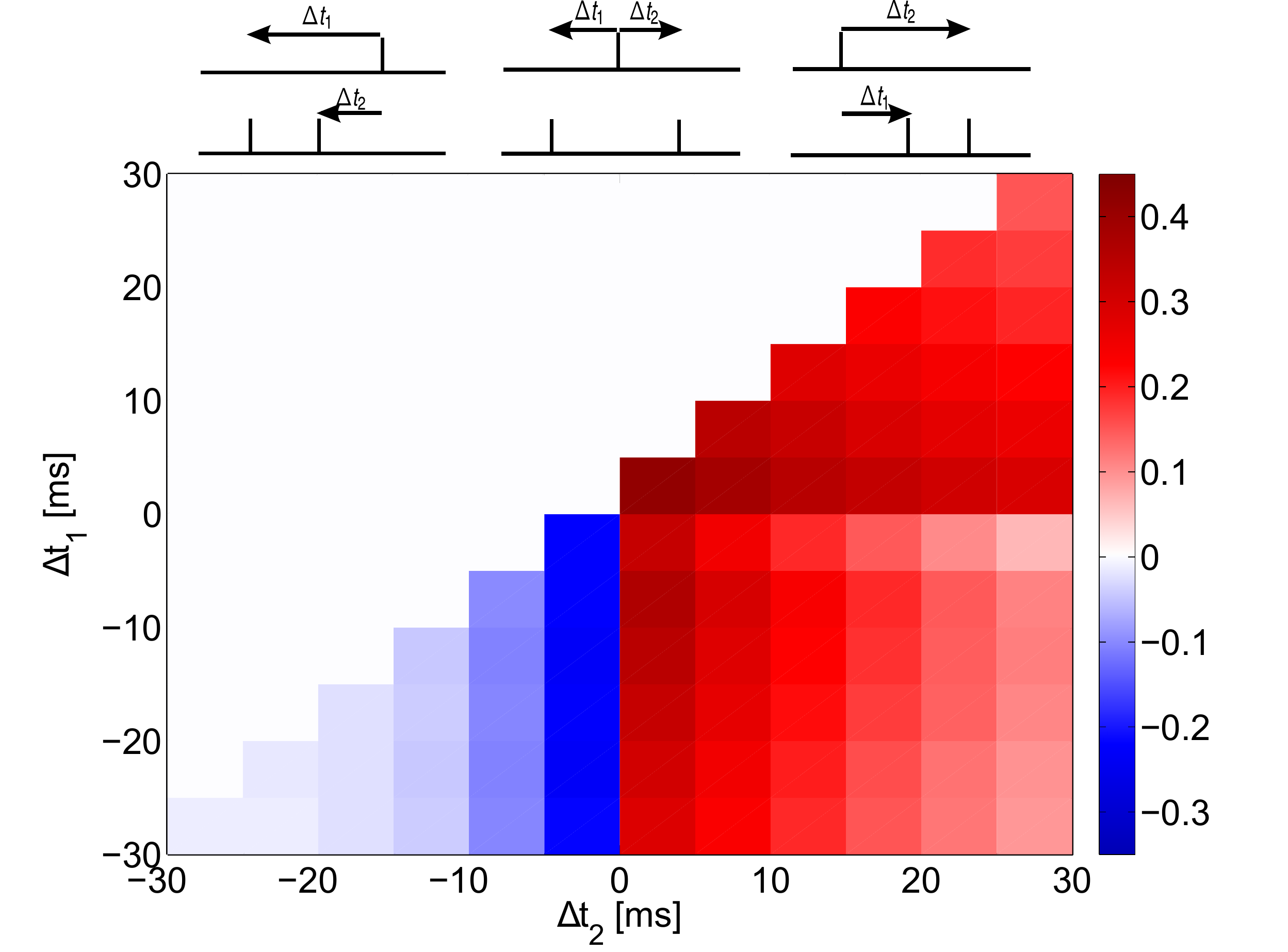}
  \caption{Synaptic weight changes in result of a suppressive triplet protocol for pre-post-post (top right triangle), post-post-pre (bottom left triangle) and post-pre-post (right bottom square) combination of spikes produced by the proposed minimal TSTDP circuit. The required bias currents taken for the triplet circuit corresponds to the hippocampal culture data set (Table~\ref{tab:1}).}\label{fig:froemke1}
\end{figure}

\begin{figure} [!t]
\centering
  \includegraphics[width=0.5\textwidth]{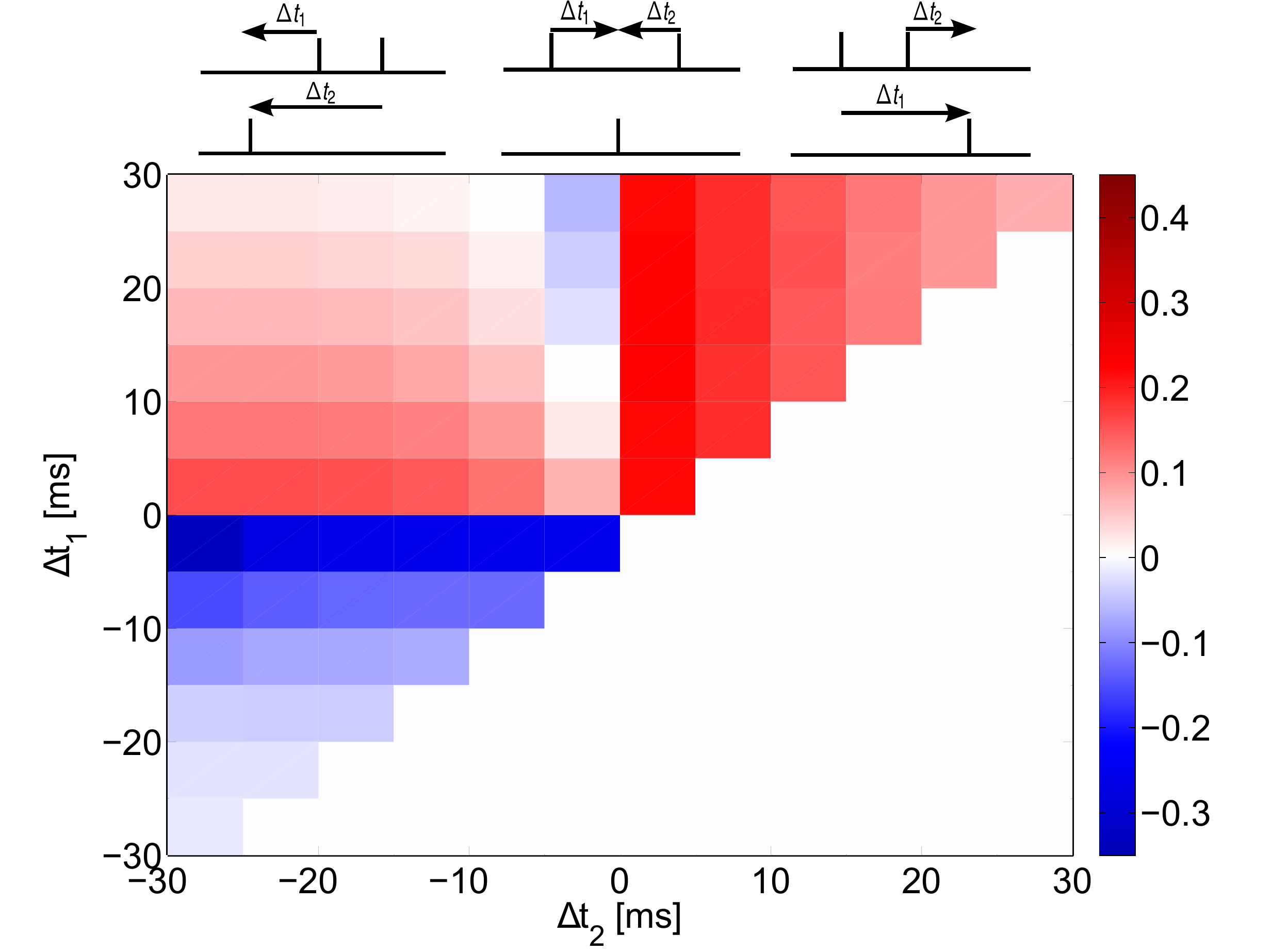}
  \caption{Synaptic weight changes as a result of a suppressive triplet protocol for pre-post-pre (top left square), pre-pre-post (top right triangle) and post-pre-pre (left bottom triangle) combination of spikes produced by the proposed minimal TSTDP circuit. The required bias currents taken for the triplet circuit corresponds to the hippocampal culture data set (Table~\ref{tab:1}).}\label{fig:froemke2}
\end{figure}

The simulation protocol (for suppressive triplet model) as described in~\cite{Froemke2002} is as follows; a third spike is added either pre- or post-synaptically to the pre-post spike pairs, to form a triplet. Then this triplet is repeated 60 times at 0.2 Hz to induce synaptic weight changes. Here, the same protocol has been used to stimulate the proposed minimal TSTDP circuit. In this protocol, there are two timing differences shown as $\Delta t_1=t_{post}-t_{pre}$ which is the timing difference between the two most left pre-post or post-pre spike pairs, and $\Delta t_2=t_{post}-t_{pre}$ which is the timing difference between the two most right pre-post or post-pre spike pairs.  
The circuit bias currents for reproducing these triplet-induced weight changes corresponds to those employed for the hippocampal data set (Table~\ref{tab:1}). 
Although the proposed circuit implements the triplet model presented in~\cite{Pfister2006} (and not the suppressive model in~\cite{Froemke2002}), obtained results showed in Figures~\ref{fig:froemke1} and~\ref{fig:froemke2} demonstrate qualitative regional agreement with the reported results in~\cite{Froemke2002}, nonetheless, there is a direct contrast between our results and their results in the post-pre-post case of spike patterns. Indeed, the triplet model weight changes induced by the pre-post-post, post-post-pre, pre-pre-post, and pre-post-post spike triplets are significantly matched to the weight changes resulted from the similar spike patterns obtained from the Froemke-Dan model. However, there is a slight difference in the results for pre-post-pre and a significant difference in the results for post-pre-post spike combinations when using these two different models.
Right bottom square in Fig.~\ref{fig:froemke1} which represents the post-pre-post case shows potentiation as it is the case for the post-pre-post spike pattern case in Fig.~\ref{fig:posprepos} also, however Froemke-Dan model results show a depression for this spike combination (Fig. 3b in~\cite{Froemke2002}). According to the discussion provided in~\cite{Pfister2006}, the difference in the result is due to the nature of the original suppressive rule where post-pre-post contributions gave rise to a depression, in contrast to TSTDP where this specific combination leads to potentiation. Note that the Froemke-Dan revised model presented in 2006 addressed this issue, since in this model there are two different potentiation and depression saturation values~\cite{Froemke2006}. This revised model now reproduces the expected experimental outcomes from \cite{Sjostrom2001}.

\begin{table*} [!t]
\centering 
\caption{Minimal TSTDP circuit bias currents and the resulted NMSEs for the two data sets}\label{tab:1}
\small \begin{tabular} {c c c c c c c c}
\hline
Data set   & $I_{\rm pot1}$ & $I_{\rm dep1}$ & $I_{\rm tp1}$ & $I_{\rm td1}$ & $I_{\rm pot2}$ & $I_{\rm tp2}$ & NMSE\\
\hline
Visual cortex      & 0  & 220~nA & 500~pA & 140~pA  & 1.15~$\mu$A  & 80~pA  & 0.33\\
\hline
Hippocampal    & 130~nA & 190~nA & 900~pA  & 170~pA  & 280~nA & 140~pA  & 1.74\\
\hline
\end{tabular}
\end{table*}

\subsection{Poissonian Protocol for the BCM Rate-based Learning}\label{subsec:bcm}

As already mentioned, in addition to the ability of reproducing the synaptic weight changes resulting from the pairing protocol (both window and change in pairing frequency), triplet protocol and quadruplet protocol (which all demonstrate the influence of timing-based variations of inputs on the synaptic weights), the proposed circuit also has the ability to give rise to a rate-based learning rule which mimics the effects of BCM. In order to demonstrate how the proposed circuit can reproduce a BCM-like behaviour, a Poissonian protocol has been used as follows. Under this protocol, the pre-synaptic and post-synaptic spike trains are generated as Poissonian spike trains with firing rate of~$\rho_{\rm pre}$ and~$\rho_{\rm post}$, respectively. This is the same protocol that has been used in~\cite{Pfister2006} to show how their proposed TSTDP model can show a close mapping to the BCM model. This paper utilizes a similar protocol to stimulate the minimal TSTDP circuit and examines if it is capable of reproducing a similar BCM-like behaviour as~\cite{Pfister2006}.

\begin{figure} [!t]
\centering
  \includegraphics[width=0.5\textwidth]{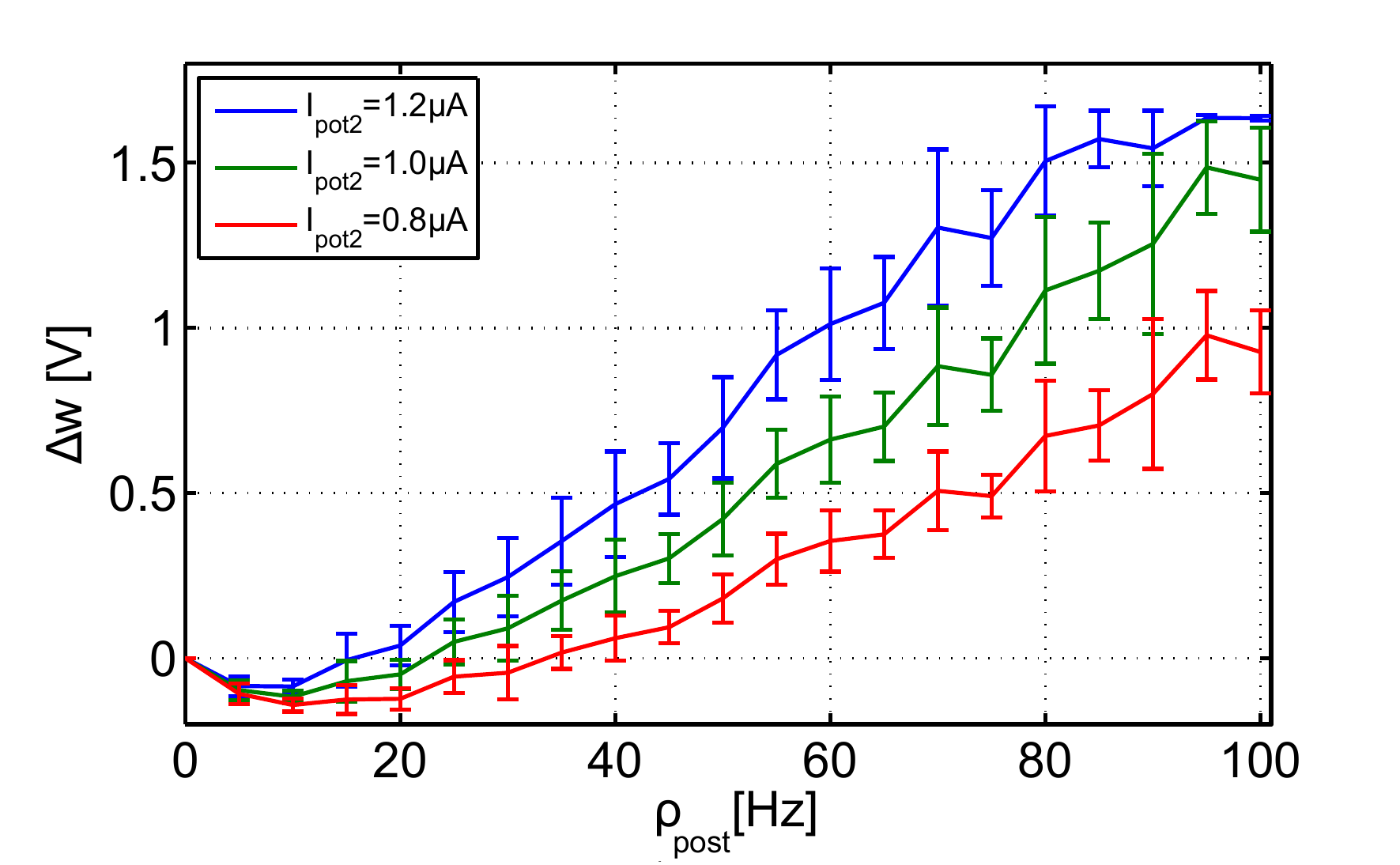}

\caption{The proposed TSTDP circuit can generate BCM-like behaviour. The required bias currents for the circuit correspond to those used for the visual cortex data set (Table~\ref{tab:1}). The three different curves show the synaptic weight changes according to three different synaptic modification thresholds demonstrating the points where LTD changes to LTP which is controlled by the current $I_{\rm pot2}$. The threshold is adjustable using the TSTDP rule parameters. In order to move the sliding threshold toward left or right, the $I_{\rm pot2}$ parameter can be altered as it is depicted in figures \ref{fig:bcm} and \ref{fig:bcm-cir}. The rate of pre-synaptic spike trains $\rho_{\rm pre}$ is 10Hz for all corresponding points. Each data point shows the mean value of the weight changes for 10 different trials and the error bars depict the standard deviations of the weight changes for each value of $\rho_{\rm post}$ for these trials.}\label{fig:bcm}
\end{figure}

In order to extract BCM-like characteristics, as described by Eq.~\ref{eq:bcm}, out of the TSTDP rule,~\cite{Pfister2006} used a minimal TSTDP rule by setting $A_3^-=0$. They specifically observed the statistical nature of the weight changes asscociated with this rule including the time averaged learning dynamics of the weight changes. Consequently, in order to show that the circuit is capable of reproducing similar BCM-like behaviour we incorporated the same protocol as used by~\cite{Pfister2006}. Therefore, either $I_{\rm dep2}$ must be set to zero in the full-triplet circuit (Fig.~\ref{fig:t-stdp}), or the circuit can be changed to the minimal TSTDP circuit presented in Fig.~\ref{fig:t-stdp1}. The simulation results for the Poissonian protocol and using the proposed minimal TSTDP circuit are shown in Fig.~\ref{fig:bcm}. 
In this figure, each data point at each post-synaptic frequency ($\rho_{\rm post}$), demonstrates the average value of weight changes for ten different realizations of post-synaptic and pre-synaptic Poissonian spike trains. In addition, each error bar shows the standard deviation of the weight changes over these ten trials. The demonstrated results were produced using the bias currents which correspond to the visual cortex data set (Table \ref{tab:1}). In the circuit, $V_{\rm post(n-1)}$, $V_{\rm post(n)}$, $\overline{V}_{\rm pre(n)}$ and ${V}_{\rm pre(n)}$ are Poissonian spike trains where~$\rho_{\rm post}$,~$\rho_{\rm post}$,~$\rho_{\rm pre}$ and~$\rho_{\rm pre}$ denote their average firing rates, respectively. The three different curves presented in Fig.~\ref{fig:bcm} display three different weight modification thresholds. In the original BCM rule, these thresholds are related to the post-synaptic firing rate, $\rho_{\rm post}$. Based on~\cite{Pfister2006}, the modification threshold for the all-to-all spike interactions can be expressed as

\begin{equation}\label{eq:tripthr}
\theta=\left\langle\rho_{\rm post}^p\right\rangle \frac{(A_2^- \tau_- A_2^+ \tau_+)}{(\rho_0^p A_3^+ \tau_+ \tau_y)},
\end{equation}
where $\left\langle\rho_{\rm post}^p\right\rangle$ is the expectation over the statistics of the $p^{\rm th}$ power of the post-synaptic firing rate and $\rho_0^p=\left\langle\rho_{\rm post}^p\right\rangle$ for large time constants (10 min or more). However, for the nearest-spike model which is the case for the proposed TSTDP circuit, it is not possible to derive a closed form expression for the modification threshold based on $\rho_{\rm post}^p$, however for post-synaptic firing rate up to 100 Hz, a similar behaviour to what Eq.~\ref{eq:tripthr} presents is inferable from the simulation results (supplementary materials of~\cite{Pfister2006}). The three different curves in Fig.~\ref{fig:bcm} are the results of three different values for $I_{\rm pot2}$ currents which correspond to three different values of $A^+_3$. This simulation suggests that the proposed circuit can not only reproduce timing-based experimental outcomes, but also can reproduce some rate-based synaptic weight modifications.

Other examples of postsynaptically driven BCM-like behaviour can be found in the supplementary materials; for these the circuit simulations were conducted by fixing the presynaptic rates to 5 Hz and 15 Hz, respectively and postsynaptic rates varied from 0 to 50 Hz. For both these cases a BCM-like behaviour was observed.

To analyze how BCM-like behaviour emerges from TSTDP, we need to go through the same analysis used by \cite{Pfister2006}.
In this circumstance, the triplet learning rule can be recast into a simpler form by considering the statistical properties of TSTDP weight changes which leads to the following time averaged equation,
\begin{eqnarray}
\left\langle \frac{dw}{dt}\right\rangle & = & -A_2^-\tau_-\rho_{\rm pre}\rho_{\rm post}+A_2^+\tau_+\rho_{\rm pre}\rho_{\rm post} \nonumber \\
& & -A_3^-\tau_-\tau_x\rho_{\rm pre}^2\rho_{\rm post}+A_3^+\tau_+\tau_y\rho_{\rm post}^2\rho_{\rm pre}, \nonumber \\
& & \label{eq:mw1}
\end{eqnarray}
where $\rho_{\rm pre}$ and $\rho_{\rm post}$ are the pre- and post-synaptic firing rates, respectively. The other parameters in the above equation $\tau_-$, and $\tau_+$, are time constants for the pair-based contribution and $\tau_x$, and $\tau_y$ are the corresponding time constants for the triplet-based contribution of the original triplet learning rule by \cite{Pfister2006}.

By considering the mapping of Eq. (\ref{eq:mw1}) into a mathematically similar functional form as Eq. (\ref{eq:bcm}), (following the method as described in~\cite{Pfister2006}) one can simply set $A_3^-=0$ and for simplicity, keep $A_2^-$ and $A_2^+$ constant in Eq. (\ref{eq:mw1}). This gives rise to the following expression 
\begin{eqnarray}
\left\langle \frac{dw}{dt}\right\rangle & = & -A_2^-\tau_-\rho_{\rm pre}\rho_{\rm post}+A_2^+\tau_+\rho_{\rm pre}\rho_{\rm post} \nonumber \\
& & \hspace{1cm}+A_3^+\tau_+\tau_y\rho_{\rm post}^2\rho_{\rm pre}. \label{eq:mw2}
\end{eqnarray}
The above equation, given an appropriate choice of parameter values, can mimic BCM-like nonlinear weight change dynamics by keeping $\rho_{\rm pre}$ fixed and altering the value of the $\rho_{\rm post}$; under these conditions, one can numerically illustrate that the weight changes as a function of increasing postsynaptic frequency, has a similar profile to the weight changes of the original BCM rule as described by Eq. (\ref{eq:bcm}).

However, one should keep in mind an important aspect of the original BCM experiments \cite{Kirkwood1996,Cooper2004} in order not to introduce any misconceptions about the original BCM rule. This aspect (excluding neuromodulatory effects) is that the original experiments were conducted using increasing presynaptic frequency of inputs \cite{Kirkwood1996}. It is a well-known and undisputed fact that neurophysiological experiments have shown that presynaptic activity typically drives postsynaptic responses, and changes in postsynaptic firing rate only occurs as a result of changes to input activity. Put simply, changes in postsynaptic firing cannot be considered independent from changes in presynaptic activity, they are functionally related. Hence, in a more precise physiological terms, the firing rate of the postsynaptic neuron really needs to be considered as a function of its presynaptic inputs. A more informative analysis of the weight dynamics of the triplet rule should take this fact about pre- and postsynaptic firing rate, i.e. $\rho_{\rm post}=F(\rho_{\rm pre})$, into account. Hence changing the postsynaptic firing rates should really be driven by changes in presynaptic firing rates, as they do in any neurophysiological setting; in this manner one can deduce a more informative link between the plasticity model and the original BCM rule. Changing $\rho_{\rm post}$ while keeping the presynaptic firing rate $\rho_{\rm pre}$ fixed, needs to be viewed with caution as it represents a misinterpretation in the application of the original stimulus protocol used in LTD/LTP experiment, despite leading to BCM-like weight changes. 

\begin{figure}[!t]
\centering
\includegraphics[width=0.5\textwidth]{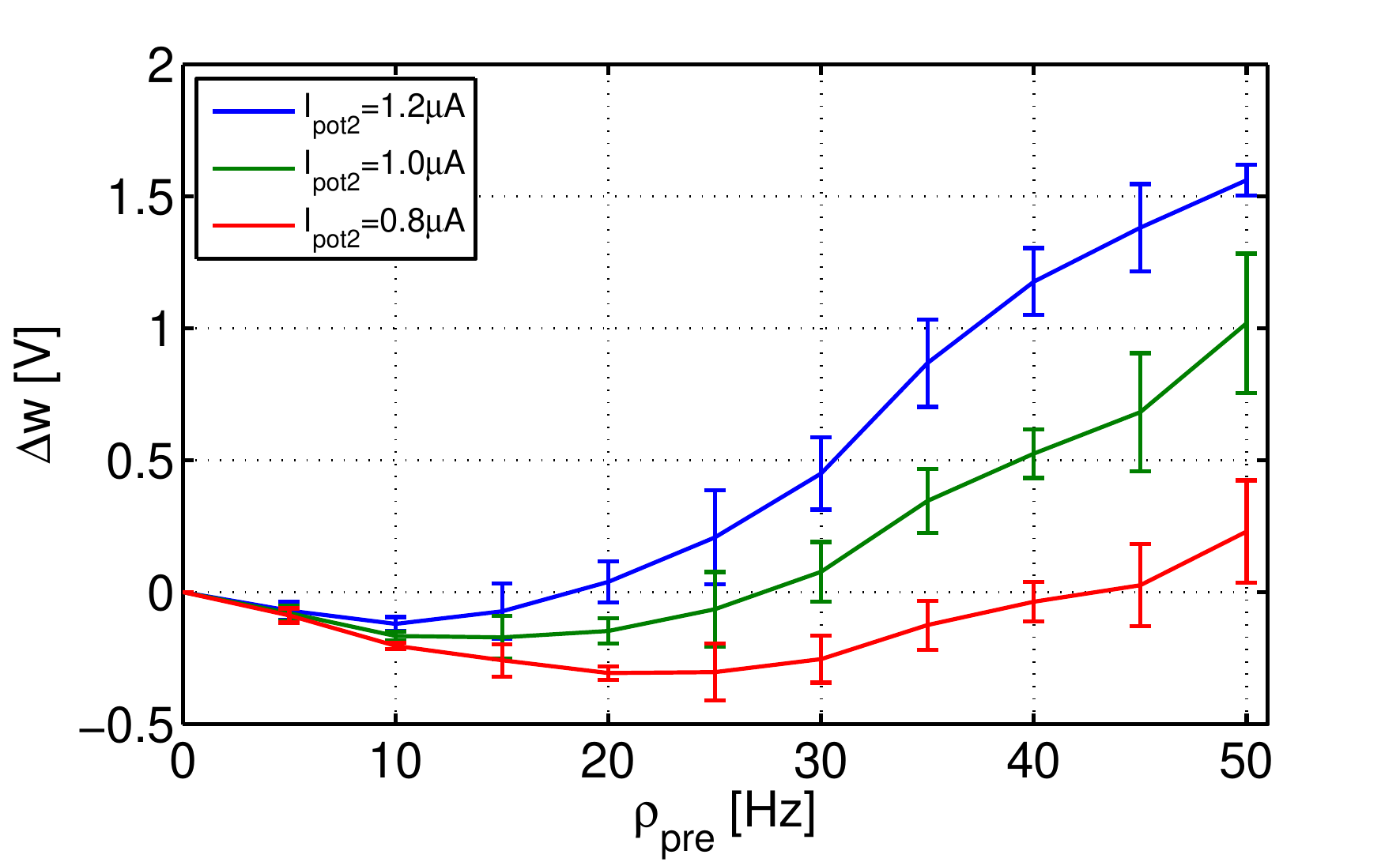}%
\caption{The proposed TSTDP circuit can generate presynaptically driven BCM-like weight changes.}%
\label{fig:bcm-cir}%
\end{figure}

As a check that our circuit could reproduce BCM-like behaviour which is driven by presynaptic (rather than postsynaptic) activity, we have repeated our circuit simulations but made the naive assumption that postsynaptic firing rate is a linear function of the presynaptic firing rate, i.e. $\rho_{\rm post}=A\rho_{\rm pre}$ and for the sake of simplicity we let $A=1$, i.e $\rho_{\rm post}=\rho_{\rm pre}$. Despite such a crude approximation, the circuit successfully was able to mimic BCM-like behaviour where weight changes were presynaptically driven, as illustrated in Fig. (\ref{fig:bcm-cir}). In this figure, each data point shows the mean value of the weight changes for 10 different trials and the error bars depict the standard deviations of the associated weight changes. 

Additionally, Matlab simulations were conducted using both the Linear Poisson neuron model and the Izhikevich model, in order to assess whether such models can reproduce presynaptically driven BCM-like changes to synaptic strength.~We found that in the case of increasing the presynaptic activity, the resulting synaptic weight changes followed a BCM-like profile where for low presynaptic activity, there was no alteration to synaptic weight; for moderate levels of presynaptic activity, gave rise to depression (LTD) and for further increases in (presynaptic) activity led to potentiation (LTP). Such a presynaptically driven BCM-like profile of synaptic change occurs for each above stated neuron model and the results of these simulations are presented in the supplementary materials. These preliminary Matlab simulations were pursued in order to inform us whether combining a circuit based model of a neuron with our TSTDP circuit will lead to a circuit implementation capable of both timing and rate based plasticity changes; this represents a future direction of our research whose results will be the subject of a future publication.

\section{Mismatch and Variation}~\label{mis}
Neuromorphic models are an approximation to biological experiments and as we can see from these experiments there is a significant variation associated with them. Nonetheless, it is of interest to produce these circuits that mimic these models, the most important usually being the trend of the circuit behavior. Having said that, the variation and mismatch inherent in the fabrication process of transistors in submicron scales and subthreshold design regime are major concern when designing analog CMOS neuromorphic circuits especially in large-scale. The majority of neuromorphic models are designed in the subthreshold regime to gain the required neuronal behavior and at the same time enjoy less power consumption compared to above threshold operational region. However, the subthreshold regime usually brings about severe transistor threshold voltage variations as well as inevitable transistor mismatches~\cite{Poon2011}. In order to minimize the effect of variations and mismatches, the analog VLSI signal processing guidelines proposed by~\cite{Vittoz1985} can be adopted. It should be acknowledged that a complete elimination of these mismatches and variations is not possible.

The proposed circuit uses a number of transistors operating in the subthreshold region and also includes current mirrors. Therefore, it is expected that this circuit will be susceptible to process variations. In order to show that the proposed design is process tolerant, two types of analysis were performed in this paper. First, the proposed design was simulated using the worst case process corners of the AMS 0.35 $\mu$m CMOS model.
The simulation results shown in Figures~\ref{fig:window} to~\ref{fig:quad} demonstrate that under the process corners the proposed circuit functions within expectation (reasonable bounds) and can show the expected behaviour in all cases. These figures show that there are slight variations in the amplitudes, which can be adjusted by retuning the circuit's bias currents. This robustness suggests that the physical implementation of the proposed design would be also robust and work within the expected design boundaries (chapter 4 of~\cite{Weste2005}).

Furthermore, since the proposed design utilizes current mirrors to copy currents and set the required amplitudes and time constants of the TSTDP model, the effect of transistors mismatch on the circuit performance must be considered. Therefore as the second variation analysis, the proposed circuit was examined against process variation and transistor mismatch. For this purpose, a 1000 Monte Carlo (MC) runs were performed on the proposed circuit in order to test its operational robustness.

The process variation scenario is as follows. All the circuit transistors go under a local variation which changes their absolute parameter values in the typical model. The process parameter that was chosen to go under these variations is the transistor threshold voltage, which is one of the most important process parameters specially in the proposed design consisting of transistors operating in the subthreshold region of operation~\cite{Seebacher2005}. The parameters vary according to the AMS C35 MC process statistical parameters. The threshold voltages of PMOS and NMOS transistors varied according to a one sigma Gaussian distribution, which can change the threshold voltages by up to 30 mV. 
Under such a circumstance, a 1000 MC runs were performed on the proposed circuit for three different cases, as described below.

\begin{figure} [!t]
\centering
  \includegraphics[width=0.51\textwidth]{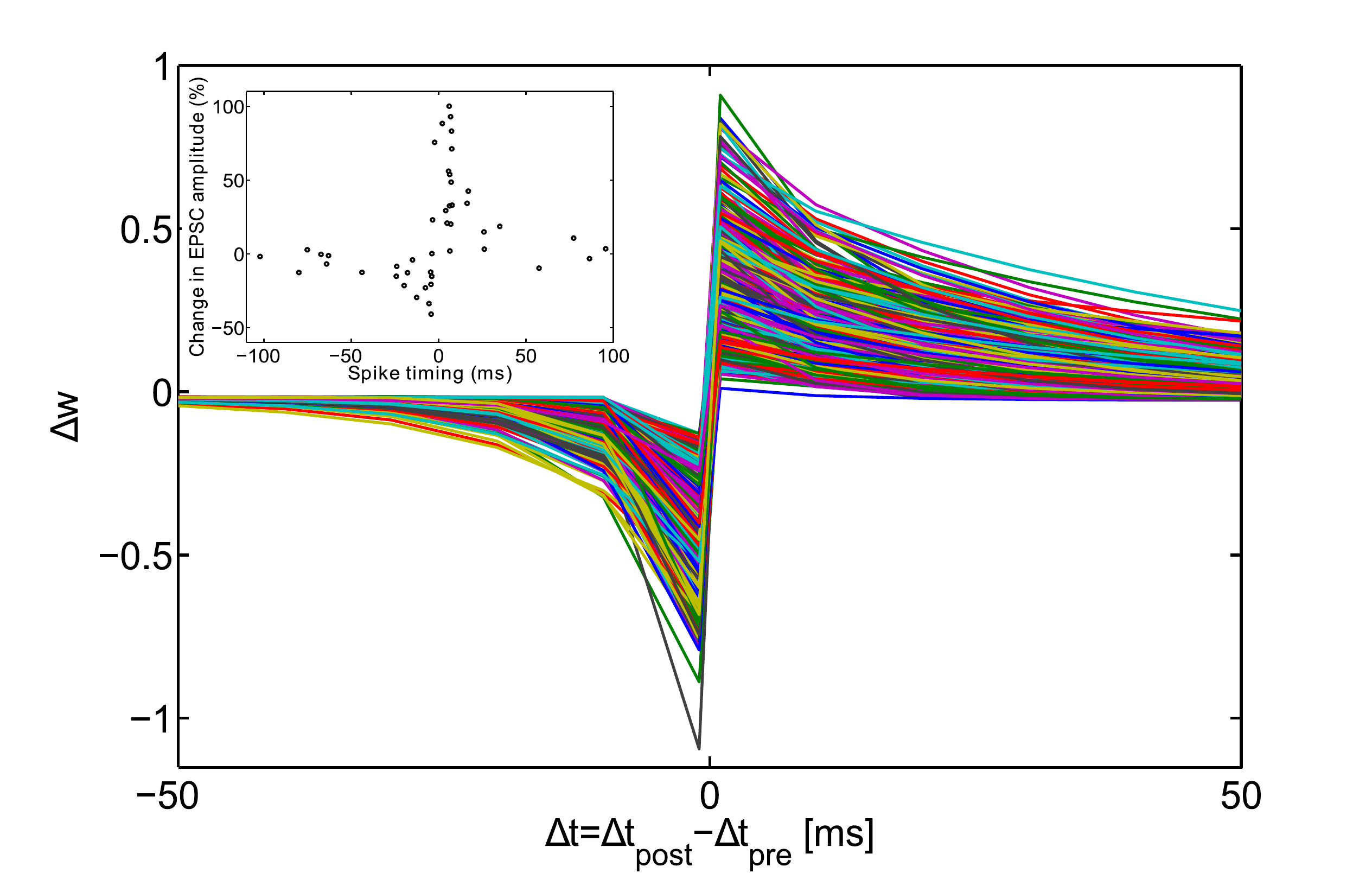}
  \caption{The proposed circuit simulation results for 1000 Monte Carlo runs for variation analysis. Each curve represents the weight change for a random set of variations on the threshold voltage of all transistors. The inset figure is the experimental data extracted from \cite{Bi1998}. Note the similarity between the simulation results and the experimental data.}\label{fig:winmc}
\end{figure}

As the first case of MC analysis, the circuit was stimulated by a pairing protocol to reproduce the exponential STDP learning window in the presence of the mentioned local variation. The circuit bias currents correspond to those used for the typical model and hippocampal data set reported in Table~\ref{tab:1}. Note that Fig.~\ref{fig:winmc} shows a 1000 MC analysis performed on the proposed circuit. This figure also shows that the proposed circuit is less susceptible to process variation and the overall LTP/LTD exponential behaviour can be preserved. However, the strength of the proposed circuit is in its controllability using the bias currents. The observed variations in the design can be alleviated by means of readjusting the circuit bias currents. This tuning can be conducted even after the circuit is fabricated. If the fabricated circuit performance changes from the expected characteristics, the circuit bias currents, which serve as inputs to the fabricated chip, can be retuned to reach the required behaviour.

\begin{figure}[!t]
\centering
\includegraphics[width=0.50\textwidth]{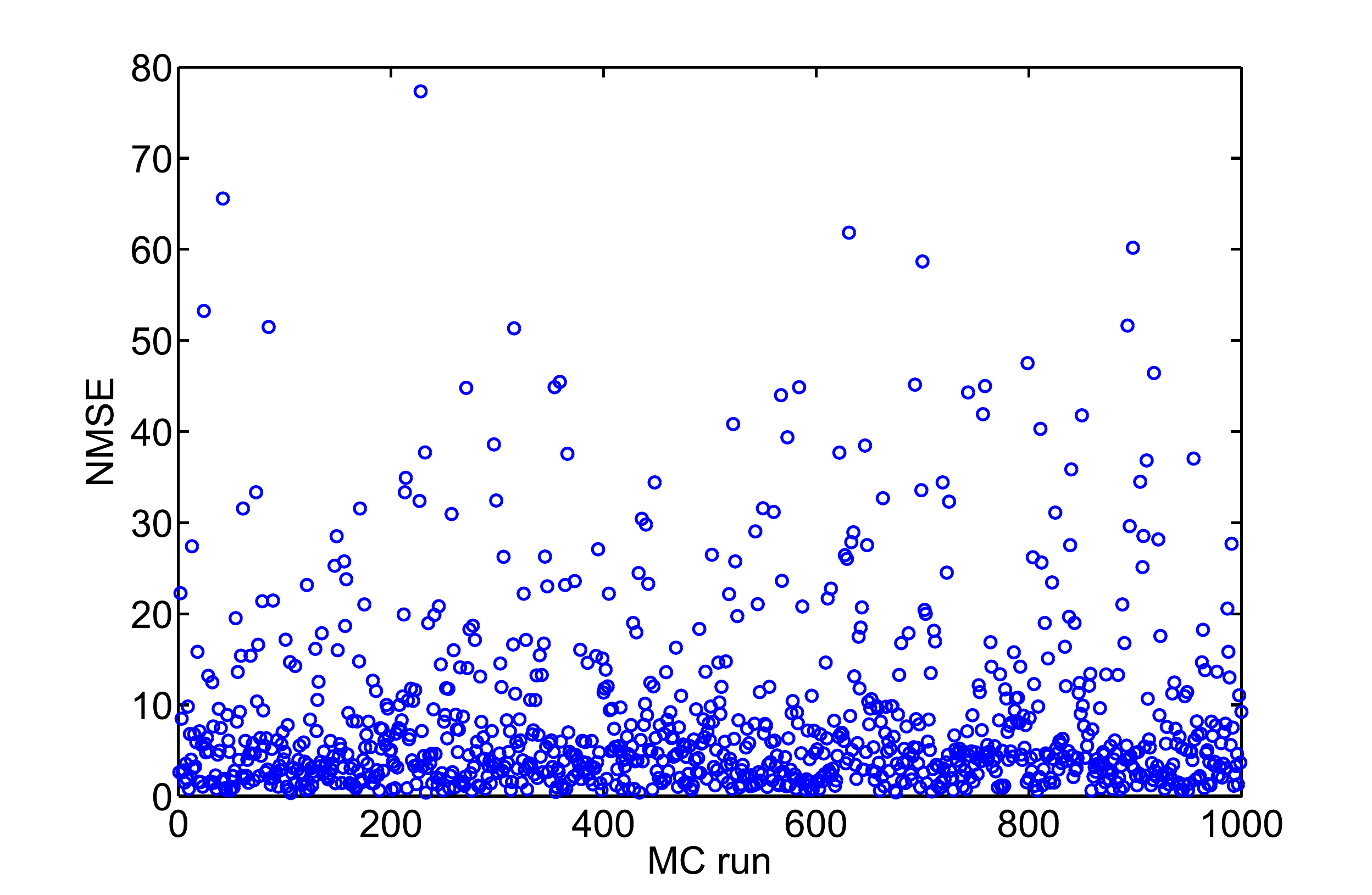}
  \caption{The proposed circuit simulation results for 1000 Monte Carlo runs for variation analysis. Each run presents a NMSE value obtained from the TSTDP circuit when employing the visual cortex data set and bias parameters (Table \ref{tab:1}), for a random variation of transistors threshold voltages.}\label{fig:pairfreqmc}
\end{figure}

\begin{figure}  [!t]
\centering
  \includegraphics[width=0.5\textwidth]{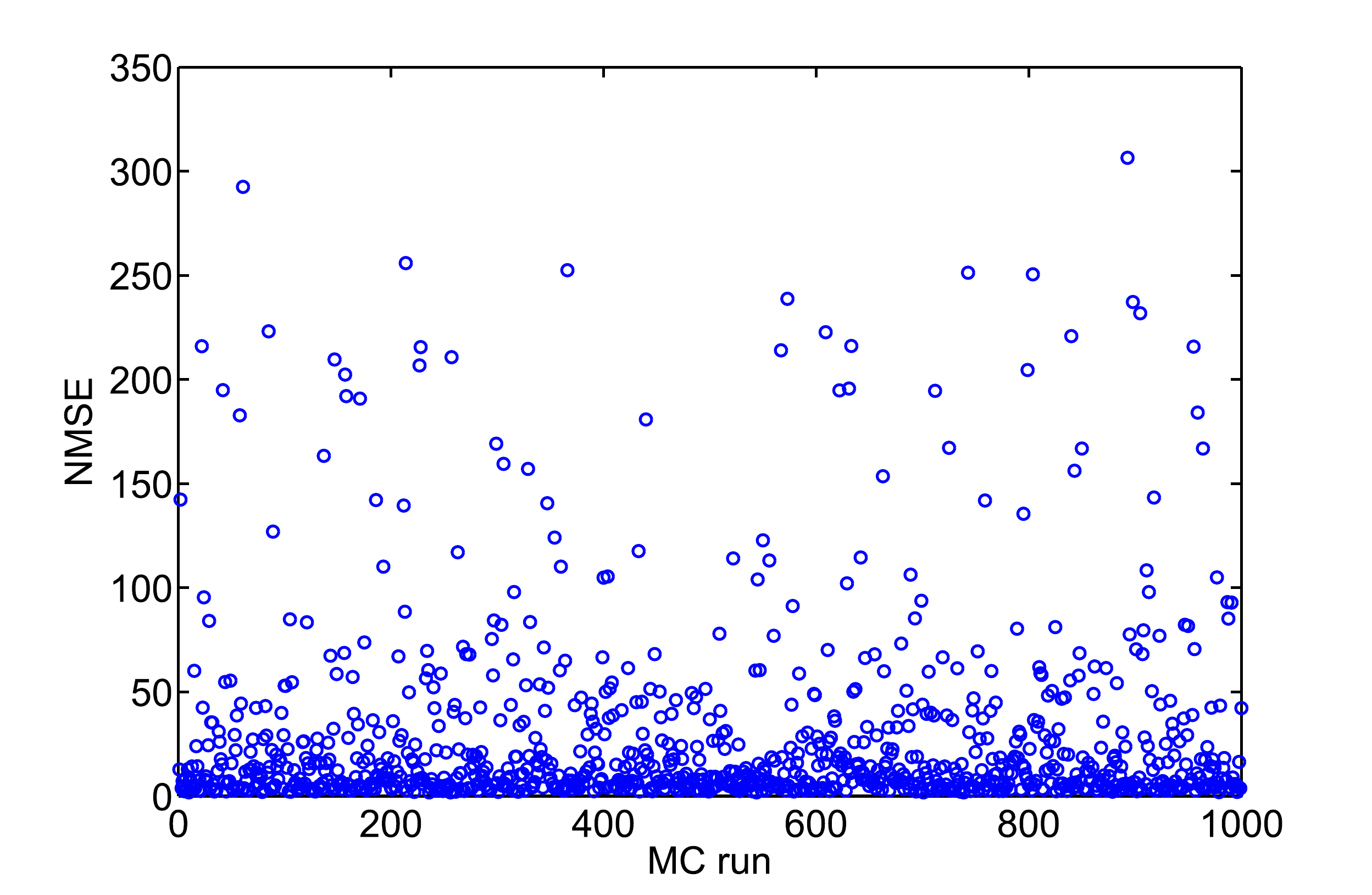}
  \caption{The proposed circuit simulation results for 1000 Monte Carlo runs for variation analysis. Each run presents a NMSE value obtained from the TSTDP circuit when employing the hippocampal data set and bias parameters (Table \ref{tab:1}), for a random variation of transistors threshold voltages.}\label{fig:hippmc}
\end{figure}

The second analysis was performed under similar process variation conditions to the first case, but this time the circuit was stimulated by the pairing protocol to reproduce the visual cortex data set and the resulting NMSEs were computed for 1000 MC runs.~The circuit bias currents correspond to those used for the typical model which are reported in Table~\ref{tab:1}.~The obtained results are shown in Fig.~\ref{fig:pairfreqmc}. Furthermore, as the third case, the same analysis was carried out for the hippocampal data set and bias parameters presented in Table~\ref{tab:1}.~Achieved results are demonstrated in Fig.~\ref{fig:hippmc}. Figures~\ref{fig:pairfreqmc} and~\ref{fig:hippmc} show significant variations in the value of NMSE compared to the typical transistor parameters that the circuit bias currents (see Table~\ref{tab:1}) were optimized for. Despite these deviations in the NMSE values under process variations, they are easily treatable by retuning the bias currents.

In the case of the visual cortex data set, the worst NMSE was almost 78 that is much larger than a minimal NMSE obtained using the typical model, ${\rm NMSE} = 0.33$. Also, in the case of hippocampal data set, the worst NMSE is about 306, which is again significantly bigger than the NMSE obtained using the typical model,~${\rm NMSE} = 1.74$. These major deviations can be significantly reduced by retuning circuit bias currents and optimizing them to get a minimal NMSE, in the presence of process variation. It means that, some bias tuning should be performed on the circuit to reach a minimal NMSE comparable to the design target.
As an example, the worst case of NMSE for the hippocampal data set (${\rm NMSE} = 306.4$) is in the case of some big changes in the threshold voltages around 30 mV. In the presence of these parameter variations in the design, all circuit bias currents were adjusted again and a new minimum NMSE was obtained.
The achieved NMSE, which is equal to 1.92, is consistent with the design expectations. The retuned circuit bias currents in this case are given in Table \ref{tab:2}. Using these new bias currents, the required behaviour for the pairing experiment (Fig.~\ref{fig:window}), the quadruplet experiment (Fig.~\ref{fig:quad}), as well as the triplet experiment (Figures~\ref{fig:prepospre} and~\ref{fig:posprepos}) were well observed.~The same approach was considered for the visual cortex data set, and the worst NMSE($=78$) changed to an acceptable ${\rm NMSE} = 0.47$ that can faithfully represent the required frequency-dependent behaviour in the pairing visual cortex experiment shown in Fig.~\ref{fig:pairfreq}. 

\begin{table*} [!t]
\centering 
\caption{Retuned TSTDP circuit bias currents and the resulted NMSEs in the presence of the worst case variation in 1000 MC runs shown in Figures~\ref{fig:pairfreqmc} and~\ref{fig:hippmc}. NMSEs were equal to 78 and 306.4 for the visual cortex and the hippocampal data sets, respectively, but they were brought back to the shown NMSEs by readjusting the circuit bias currents from the values shown in Table~\ref{tab:1}.}\label{tab:2}
\small \begin{tabular} {c c c c c c c c}
\hline
Data set   & $I_{\rm pot1}$ & $I_{\rm dep1}$ & $I_{\rm tp1}$ & $I_{\rm td1}$ & $I_{\rm pot2}$ & $I_{\rm tp2}$ & NMSE\\
\hline
Visual cortex    & 0  & 260~nA & 1~nA & 120~pA  & 590~nA  & 150~pA  & 0.47\\
\hline
Hippocampal      & 510~nA & 240~nA & 270~pA  & 860~pA  & 110~nA & 180~pA  & 1.92\\
\hline
\end{tabular}
\end{table*} 

Both worst case and MC analysis performed on the circuit show the robustness and the controllability of the design in the presence of physical variations. Hence, despite the fact that the proposed design has some susceptibility to process variations, a post-fabrication calibration is possible through retuning the bias currents of the design to achieve a minimal NMSE to faithfully reproduce the needed learning behaviour.

Although the presented MC simulation analysis and the result from fine-tuned circuits show that it is possible to minimize the effect of process variations, fine tuning of the circuit bias parameters in a large spiking neural network of neurons and proposed triplet synapses sounds to be not practical. Hence a variation aware design technique is required that take into account the process variation while designing the desired neuronal circuit. The technique utilized to alleviate variations and mismatch in our design is using the rules of transistor matching proposed by~\cite{Vittoz1985}. Another approach available in the literature, which unfortunately is not useful for our proposed circuit design, is the design technique proposed and well utilized by~\cite{Rachmuth2011,Meng2011} in neuromorphic modeling of ion channel and ionic dynamic in large scale neuronal networks. This variation aware design technique exploits source degeneration and negative feedback methods to increase the dynamic range of input voltages of transistors and make them robust against mismatch errors that happen majorly because of the low input voltage dynamic range in traditional subthreshold current mode circuits~\cite{Poon2011}.

A direction for future research is to use the source degeneration and negative feedback design technique and build a network of neurons with TSTDP synapses which is capable of pattern selection as described in~\cite{Gjorgjieva2011}.      

\section{Discussion and Conclusion}
\label{conc}

The development of biologically inspired chips has had a long history since the work of~\cite{Mead1989} and is currently seen as an area of increasing activity. The development of these chips based upon the dynamics of the brain has many applications ranging from smart sensor for collision avoidance through to self autonomous robotic systems. Naturally an important step is to explore and further develop current/new technologies where the operation, and where necessary, the underlying mechanisms found in the brain can be translated from a biological setting to {\it in-silico} chips. This translation has many unresolved issues, including that of packing density and adaptive connectivity between the processing centres of the chip. Recent works have focused on how to increase the packing density of simple spiking neurons, where the integrate-and-fire neuron model is typically used in VLSI implementations due to its low area and energy implementation costs. In contrast there have been fewer attempts at implementing synapses, especially ones which are capable of altering their respective responses in an activity-dependent fashion.

Recent attempts have investigated different design strategies for synapses which change according to some synaptic plasticity rules e.g.~\cite{Bofill-I-Petit2004,Indiveri2006,Mayr2010a,Meng2011,Rachmuth2011}. Most previous attempts have focused on implementations of PSTDP, where the implemented circuits demonstrate the trend presented by their rules~\cite{Bamford2012}. Nonetheless, they tend to present different quantitative properties. \cite{Bofill-I-Petit2004} have presented a PSTDP circuit which can faithfully reproduce the exponential decay profiles typically presented in computational studies. In a more recent work, the symmetric PSTDP implementation by \cite{Indiveri2006} was shown to reproduce a plasticity window whose temporal profile was qualitatively similar to those previously observed in experiments by \cite{Bi1998}, but could not capture the exponential decays present in the main PSTDP model (Eq. \ref{eq:stdppair}).

In parallel to the above mentioned studies, different approaches to implementing plastic synapses in VLSI have been pursued, of interest are the ones which have designed circuits where synaptic change is jointly determined by several variables of physiological significance. For example the work by \cite{Mayr2010} proposed a new circuit design which incorporated known synaptic state variables and was capable to qualitatively reproduce the outcomes of rate-based and also some spike-based synaptic plasticity experiments. Notably, following this philosophy, there has been a concerted effort to mimic the biophysical nature of synapses, by implementing biophysically-inspired VLSI circuit designs, where the recent work by \cite{Meng2011,Rachmuth2011} have presented a VLSI implementation of a spiking neuron and synapses. Following known biophysical processes, the synapse circuit mimics the multiscale temporal nature of voltage and calcium evolution, consequently giving rise to the well known and documented BCM synaptic plasticity rule (\cite{Bienenstock1982}). 

Our recent work complements this research, where we have extended previous principles towards PSTDP circuit design and took into consideration higher order spike contributions/interactions and implemented circuits capable of TSTDP. Here, a new triplet-based Spike Timing Dependent Plasticity circuit was proposed. The circuit was examined against various patterns of spikes and experimentally used induction protocols, from a spike pairing protocol to a Poissonian spiking protocol. It was shown that the proposed design is capable of demonstrating the exponential learning window of the STDP~\cite{Bi1998}, the pairing frequency effect on the synaptic weight changes~\cite{Sjostrom2001}, the weight changes induced by a quadruplet experiments, as well as an appropriate behaviour to six different patterns of spike triplets~\cite{Pfister2006,Froemke2002,Wang2005}. In addition, it was demonstrated that the proposed timing based circuit can give rise (as a first order approximation) to BCM-like behaviour which is a rate-based synaptic plasticity rule. To the best of our knowledge, this is the first VLSI circuit which is capable of reproducing all mentioned types of experiments.
The proposed circuit, hence, can be utilized in the implementation of a VLSI synapse which possesses timing- and rate-based plasticity features, simultaneously. This synapse then can be employed in various neuromorphic systems aiming for different applications ranging from classifying complex patterns~\cite{Mitra2009}, to a TSTDP circuit which generalizes the BCM rule to higher order spatio-temporal correlations~\cite{Gjorgjieva2011}. 
All These features make the proposed circuit a valued and interesting contribution to the silicon based synaptic neural systems and pave the way for a realistic neuromorphic engineering implementation.

\section*{Acknowledgment}
The support of the Australian Research Council (ARC) is gratefully acknowledged.~Authors would also like to thank Dr Jean-Pascal Pfister and Prof Wulfram Gerstner for useful communications.

\bibliographystyle{plainnat}
\bibliography{Refs}

\begin{thebibliography}{42}
\expandafter\ifx\csname natexlab\endcsname\relax\def\natexlab#1{#1}\fi
\expandafter\ifx\csname url\endcsname\relax
  \def\url#1{{\tt #1}}\fi

\bibitem[Azghadi et~al.(2012{\natexlab{a}})Azghadi, Al-Sarawi, Iannella, and
  Abbott]{RahimiAzghadi2012a}
M.~Rahimi Azghadi, S.~Al-Sarawi, N.~Iannella, and D.~Abbott.
\newblock Design and implementation of {BCM} rule based on spike-timing
  dependent plasticity.
\newblock In {\em The 2012 International Joint Conference on Neural Networks
  (IJCNN)}, pages 1--7. IEEE, 2012{\natexlab{a}}.

\bibitem[Azghadi et~al.(2012{\natexlab{b}})Azghadi, Al-Sarawi, Iannella, and
  Abbott]{RahimiAzghadi2012}
M.~Rahimi Azghadi, S.~Al-Sarawi, N.~Iannella, and D.~Abbott.
\newblock Efficient design of triplet based spike-timing dependent plasticity.
\newblock In {\em The 2012 International Joint Conference on Neural Networks
  (IJCNN)}, pages 1--7. IEEE, 2012{\natexlab{b}}.

\bibitem[Azghadi et~al.(2011)Azghadi, Kavehei, Al-Sarawi, Iannella, and
  Abbott]{RahimiAzghadi2011}
M.~Rahimi Azghadi, O.~Kavehei, S.~Al-Sarawi, N.~Iannella, and D.~Abbott.
\newblock Novel {VLSI} implementation for triplet-based spike-timing dependent
  plasticity.
\newblock In {\em Proceedings of the 7th International Conference on
  Intelligent Sensors, Sensor Networks and Information Processing}, pages
  158--162, 2011.

\bibitem[Bamford et~al.(2012)Bamford, Murray, and Willshaw]{Bamford2012}
S.~Bamford, A.~Murray, and D.~Willshaw.
\newblock Spike-timing dependent plasticity with weight dependence evoked from
  physical constraints.
\newblock {\em IEEE Transactions on Biomedical Circuits and Systems},
  6\penalty0 (4):\penalty0 385--398, 2012.

\bibitem[Bi and Poo(1998)]{Bi1998}
G.~Bi and M.~Poo.
\newblock Synaptic modifications in cultured hippocampal neurons: dependence on
  spike timing, synaptic strength, and postsynaptic cell type.
\newblock {\em The Journal of Neuroscience}, 18\penalty0 (24):\penalty0
  10464--10472, 1998.

\bibitem[Bienenstock et~al.(1982)Bienenstock, Cooper, and
  Munro]{Bienenstock1982}
E.L. Bienenstock, L.N. Cooper, and P.W. Munro.
\newblock Theory for the development of neuron selectivity: orientation
  specificity and binocular interaction in visual cortex.
\newblock {\em The Journal of Neuroscience}, 2\penalty0 (1):\penalty0 32, 1982.

\bibitem[Bofill-I-Petit and Murray(2004)]{Bofill-I-Petit2004}
A.~Bofill-I-Petit and A.F. Murray.
\newblock Synchrony detection and amplification by silicon neurons with {STDP}
  synapses.
\newblock {\em IEEE transactions on neural networks/a publication of the IEEE
  Neural Networks Council}, 15\penalty0 (5):\penalty0 1296--1304, 2004.

\bibitem[Brader et~al.(2007)Brader, Senn, and Fusi]{Brader2007}
J.M. Brader, W.~Senn, and S.~Fusi.
\newblock Learning real-world stimuli in a neural network with spike-driven
  synaptic dynamics.
\newblock {\em Neural computation}, 19\penalty0 (11):\penalty0 2881--2912,
  2007.

\bibitem[Cameron et~al.(2005)Cameron, Boonsobhak, Murray, and
  Renshaw]{Cameron2005}
K.~Cameron, V.~Boonsobhak, A.~Murray, and D.~Renshaw.
\newblock Spike timing dependent plasticity ({STDP}) can ameliorate process
  variations in neuromorphic {VLSI}.
\newblock {\em IEEE Transactions on Neural Networks}, 16\penalty0 (6):\penalty0
  1626--1637, 2005.

\bibitem[Clopath and Gerstner(2010)]{Clopath2010}
C.~Clopath and W.~Gerstner.
\newblock Voltage and spike timing interact in {STDP}--a unified model.
\newblock {\em Frontiers in Synaptic Neuroscience}, 2:\penalty0 art. no.~25,
  2010.

\bibitem[Cooper et~al.(2004)Cooper, Intrator, Blais, and Shouval]{Cooper2004}
L.N. Cooper, N.~Intrator, B.S. Blais, and H.Z. Shouval.
\newblock {\em Theory of Cortical Plasticity}.
\newblock World Scientific Pub Co Inc, 2004.

\bibitem[Dayan and Abbott(2001)]{Dayan2001}
P.~Dayan and L.F. Abbott.
\newblock {\em Theoretical Neuroscience: Computational and Mathematical
  Modeling of Neural Systems}.
\newblock Taylor \& Francis, 2001.

\bibitem[Farquhar and Hasler(2005)]{Farquhar2005}
E.~Farquhar and P.~Hasler.
\newblock A bio-physically inspired silicon neuron.
\newblock {\em Circuits and Systems I: Regular Papers, IEEE Transactions on},
  52\penalty0 (3):\penalty0 477--488, 2005.

\bibitem[Froemke and Dan(2002)]{Froemke2002}
R.C. Froemke and Y.~Dan.
\newblock Spike-timing-dependent synaptic modification induced by natural spike
  trains.
\newblock {\em Nature}, 416\penalty0 (6879):\penalty0 433--438, 2002.

\bibitem[Froemke et~al.(2006)Froemke, Tsay, Raad, Long, and Dan]{Froemke2006}
R.C. Froemke, I.A. Tsay, M.~Raad, J.D. Long, and Y.~Dan.
\newblock Contribution of individual spikes in burst-induced long-term synaptic
  modification.
\newblock {\em Journal of neurophysiology}, 95\penalty0 (3):\penalty0
  1620--1629, 2006.

\bibitem[Gerstner et~al.(1996)Gerstner, Kempter, Van~Hemmen, and
  Wagner]{Gerstner1996}
W.~Gerstner, R.~Kempter, J.L. Van~Hemmen, and H.~Wagner.
\newblock A neuronal learning rule for sub-millisecond temporal coding.
\newblock {\em Nature}, 383\penalty0 (6595):\penalty0 76--78, 1996.

\bibitem[Gjorgjieva et~al.(2011)Gjorgjieva, Clopath, Audet, and
  Pfister]{Gjorgjieva2011}
J.~Gjorgjieva, C.~Clopath, J.~Audet, and J.P. Pfister.
\newblock A triplet spike-timing--dependent plasticity model generalizes the
  bienenstock--cooper--munro rule to higher-order spatiotemporal correlations.
\newblock {\em Proceedings of the National Academy of Sciences}, 108\penalty0
  (48):\penalty0 19383--19388, 2011.

\bibitem[Iannella and Tanaka(2006)]{Iannella2006}
N.~Iannella and S.~Tanaka.
\newblock Synaptic efficacy cluster formation across the dendrite via stdp.
\newblock {\em Neuroscience letters}, 403\penalty0 (1-2):\penalty0 24--29,
  2006.

\bibitem[Iannella et~al.(2010)Iannella, Launey, and Tanaka]{Iannella2010}
N.L. Iannella, T.~Launey, and S.~Tanaka.
\newblock Spike timing-dependent plasticity as the origin of the formation of
  clustered synaptic efficacy engrams.
\newblock {\em Frontiers in Computational Neuroscience}, 4:\penalty0 art. no.
  20., 2010.

\bibitem[Indiveri et~al.(2006)Indiveri, Chicca, and Douglas]{Indiveri2006}
G.~Indiveri, E.~Chicca, and R.~Douglas.
\newblock A {VLSI} array of low-power spiking neurons and bistable synapses
  with spike-timing dependent plasticity.
\newblock {\em IEEE Transactions on Neural Networks}, 17\penalty0 (1):\penalty0
  211--221, 2006.

\bibitem[Indiveri et~al.(2011)Indiveri, Linares-Barranco, Hamilton, Van~Schaik,
  Etienne-Cummings, Delbruck, Liu, Dudek, H{\"a}fliger, Renaud, Schemmel,
  Cauwenberghs, Arthur, Hynna, Folowosele, Saighi, Serrano-Gotarredona,
  Wijekoon, Wang, and Boahen]{Indiveri2011}
G.~Indiveri, B.~Linares-Barranco, T.J. Hamilton, A.~Van~Schaik,
  R.~Etienne-Cummings, T.~Delbruck, S.C. Liu, P.~Dudek, P.~H{\"a}fliger,
  S.~Renaud, J.~Schemmel, G.~Cauwenberghs, J.~Arthur, K.~Hynna, F.~Folowosele,
  S.~Saighi, T.~Serrano-Gotarredona, J.~Wijekoon, Y.~Wang, and K.~Boahen.
\newblock Neuromorphic silicon neuron circuits.
\newblock {\em Frontiers in Neuroscience}, 5:\penalty0 art. no. 73., 2011.

\bibitem[Kirkwood et~al.(1996)Kirkwood, Rioult, and Bear]{Kirkwood1996}
A.~Kirkwood, M.G. Rioult, and M.F. Bear.
\newblock Experience-dependent modification of synaptic plasticity in visual
  cortex.
\newblock {\em Nature}, 381\penalty0 (6582):\penalty0 526--528, 1996.

\bibitem[Koickal et~al.(2007)Koickal, Hamilton, Tan, Covington, Gardner, and
  Pearce]{Koickal2007}
T.J. Koickal, A.~Hamilton, S.L. Tan, J.A. Covington, J.W. Gardner, and T.C.
  Pearce.
\newblock Analog {VLSI} circuit implementation of an adaptive neuromorphic
  olfaction chip.
\newblock {\em IEEE Transactions on Circuits and Systems I}, 54\penalty0
  (1):\penalty0 60--73, 2007.

\bibitem[Mayr et~al.(2010)Mayr, Noack, Partzsch, and Schuffny]{Mayr2010}
C.~Mayr, M.~Noack, J.~Partzsch, and R.~Schuffny.
\newblock Replicating experimental spike and rate based neural learning in
  {CMOS}.
\newblock In {\em Proceedings of IEEE International Symposium on Circuits and
  Systems (ISCAS)}, pages 105--108, 2010.

\bibitem[Mayr and Partzsch(2010)]{Mayr2010a}
C.G. Mayr and J.~Partzsch.
\newblock Rate and pulse based plasticity governed by local synaptic state
  variables.
\newblock {\em Frontiers in Synaptic Neuroscience}, 2:\penalty0 art. no. 33,
  2010.

\bibitem[Mead(1989)]{Mead1989}
C.~Mead.
\newblock {\em Analog VLSI and Neural Systems}.
\newblock Addison-Wesley, 1989.

\bibitem[Meng et~al.(2011)Meng, Zhou, Monzon, and Poon]{Meng2011}
Y.~Meng, K.~Zhou, J.J.C. Monzon, and C.S. Poon.
\newblock Iono-neuromorphic implementation of spike-timing-dependent synaptic
  plasticity.
\newblock In {\em 2011 Annual International Conference of the IEEE Engineering
  in Medicine and Biology Society, EMBC}, pages 7274--7277, 2011.

\bibitem[Mitra et~al.(2009)Mitra, Fusi, and Indiveri]{Mitra2009}
S.~Mitra, S.~Fusi, and G.~Indiveri.
\newblock Real-time classification of complex patterns using spike-based
  learning in neuromorphic {VLSI}.
\newblock {\em IEEE Transactions on Biomedical Circuits and Systems,},
  3\penalty0 (1):\penalty0 32--42, 2009.

\bibitem[Oja(1982)]{Oja1982}
E.~Oja.
\newblock Simplified neuron model as a principal component analyzer.
\newblock {\em Journal of Mathematical Biology}, 15\penalty0 (3):\penalty0
  267--273, 1982.

\bibitem[Pfister and Gerstner(2006)]{Pfister2006}
J.P. Pfister and W.~Gerstner.
\newblock Triplets of spikes in a model of spike timing-dependent plasticity.
\newblock {\em The Journal of Neuroscience}, 26\penalty0 (38):\penalty0
  9673--9682, 2006.

\bibitem[Poon and Zhou(2011)]{Poon2011}
C.S. Poon and K.~Zhou.
\newblock Neuromorphic silicon neurons and large-scale neural networks:
  challenges and opportunities.
\newblock {\em Frontiers in neuroscience}, 5, 2011.

\bibitem[Rachmuth et~al.(2011)Rachmuth, Shouval, Bear, and Poon]{Rachmuth2011}
G.~Rachmuth, H.Z. Shouval, M.F. Bear, and C.S. Poon.
\newblock A biophysically-based neuromorphic model of spike rate-and
  timing-dependent plasticity.
\newblock {\em Proceedings of the National Academy of Sciences}, 108\penalty0
  (49):\penalty0 E1266--E1274, 2011.

\bibitem[Ramakrishnan et~al.(2011)Ramakrishnan, Hasler, and
  Gordon]{Ramakrishnan2011}
S.~Ramakrishnan, P.E. Hasler, and C.~Gordon.
\newblock Floating gate synapses with spike-time-dependent plasticity.
\newblock {\em IEEE Transactions on Biomedical Circuits and Systems},
  5\penalty0 (3):\penalty0 244--252, 2011.

\bibitem[Schemmel et~al.(2006)Schemmel, Grubl, Meier, and
  Mueller]{Schemmel2006}
J.~Schemmel, A.~Grubl, K.~Meier, and E.~Mueller.
\newblock Implementing synaptic plasticity in a vlsi spiking neural network
  model.
\newblock In {\em Neural Networks, 2006. IJCNN'06. International Joint
  Conference on}, pages 1--6. IEEE, 2006.

\bibitem[Seebacher(2005)]{Seebacher2005}
E.~Seebacher.
\newblock {CMOS} technology characterisation for analog/{RF} application.
\newblock In {\em 11th Workshop on Electronics for LHC and Future Experiments},
  page~33. CERN, 2005.

\bibitem[Simoni et~al.(2004)Simoni, Cymbalyuk, Sorensen, Calabrese, and
  DeWeerth]{Simoni2004}
M.F. Simoni, G.S. Cymbalyuk, M.E. Sorensen, R.L. Calabrese, and S.P. DeWeerth.
\newblock A multiconductance silicon neuron with biologically matched dynamics.
\newblock {\em Biomedical Engineering, IEEE Transactions on}, 51\penalty0
  (2):\penalty0 342--354, 2004.

\bibitem[Sj{\"{o}}str{\"o}m et~al.(2001)Sj{\"{o}}str{\"o}m, Turrigiano, and
  Nelson]{Sjostrom2001}
P.J. Sj{\"{o}}str{\"o}m, G.G. Turrigiano, and S.B. Nelson.
\newblock Rate, timing, and cooperativity jointly determine cortical synaptic
  plasticity.
\newblock {\em Neuron}, 32\penalty0 (6):\penalty0 1149--1164, 2001.

\bibitem[Song et~al.(2000)Song, Miller, and Abbott]{Song2000}
S.~Song, K.D. Miller, and L.F. Abbott.
\newblock Competitive hebbian learning through spike-timing-dependent synaptic
  plasticity.
\newblock {\em Nature Neuroscience}, 3:\penalty0 919--926, 2000.

\bibitem[Tanaka et~al.(2009)Tanaka, Morie, and Aihara]{Tanaka2009}
H.~Tanaka, T.~Morie, and K.~Aihara.
\newblock A {CMOS} spiking neural network circuit with symmetric/asymmetric
  {STDP} function.
\newblock {\em IEICE Transactions on Fundamentals of Electronics,
  Communications and Computer Sciences}, {E92-A}\penalty0 (7):\penalty0
  1690--1698, 2009.

\bibitem[Vittoz(1985)]{Vittoz1985}
E.A. Vittoz.
\newblock The design of high-performance analog circuits on digital {CMOS}
  chips.
\newblock {\em Solid-State Circuits, IEEE Journal of}, 20\penalty0
  (3):\penalty0 657--665, 1985.

\bibitem[Wang et~al.(2005)Wang, Gerkin, Nauen, and Bi]{Wang2005}
H.X. Wang, R.C. Gerkin, D.W. Nauen, and G.Q. Bi.
\newblock Coactivation and timing-dependent integration of synaptic
  potentiation and depression.
\newblock {\em Nature Neuroscience}, 8\penalty0 (2):\penalty0 187--193, 2005.

\bibitem[Weste and Harris(2005)]{Weste2005}
N.H.E. Weste and D.~Harris.
\newblock {\em CMOS VLSI Design: A Circuits and Systems Perspective}.
\newblock Addison-Wesley, 2005.

\end{thebibliography}







\end{document}